\documentclass{article} % For LaTeX2e
\usepackage{arxiv}
\usepackage{times}

\usepackage{amsmath,amsfonts,bm}

\def\eqref#1{equation~\ref{#1}}
\def\1{\bm{1}}

\DeclareMathAlphabet{\mathsfit}{\encodingdefault}{\sfdefault}{m}{sl}
\SetMathAlphabet{\mathsfit}{bold}{\encodingdefault}{\sfdefault}{bx}{n}

\usepackage{amsmath}
\usepackage{amssymb}
\usepackage{graphicx}
\usepackage{booktabs}
\usepackage{wrapfig}
\usepackage{multirow}
\usepackage{xcolor}
\usepackage{xspace}
\usepackage{hyperref}
\usepackage{url}
\usepackage[capitalize]{cleveref}

\title{ViewStruct: View-Grounded Articulation and Connectivity Inference}

\author{Wenjing Bian$^{1}$ \quad Ranran Huang$^{2}$ \quad Andreas Geiger$^{1}$ \\[4pt]
  $^{1}$University of T\"ubingen, T\"ubingen AI Center \quad $^{2}$Imperial College London
}

\begin{document}

\maketitle

\begin{figure}[h]
    \centering
    \includegraphics[width=\textwidth]{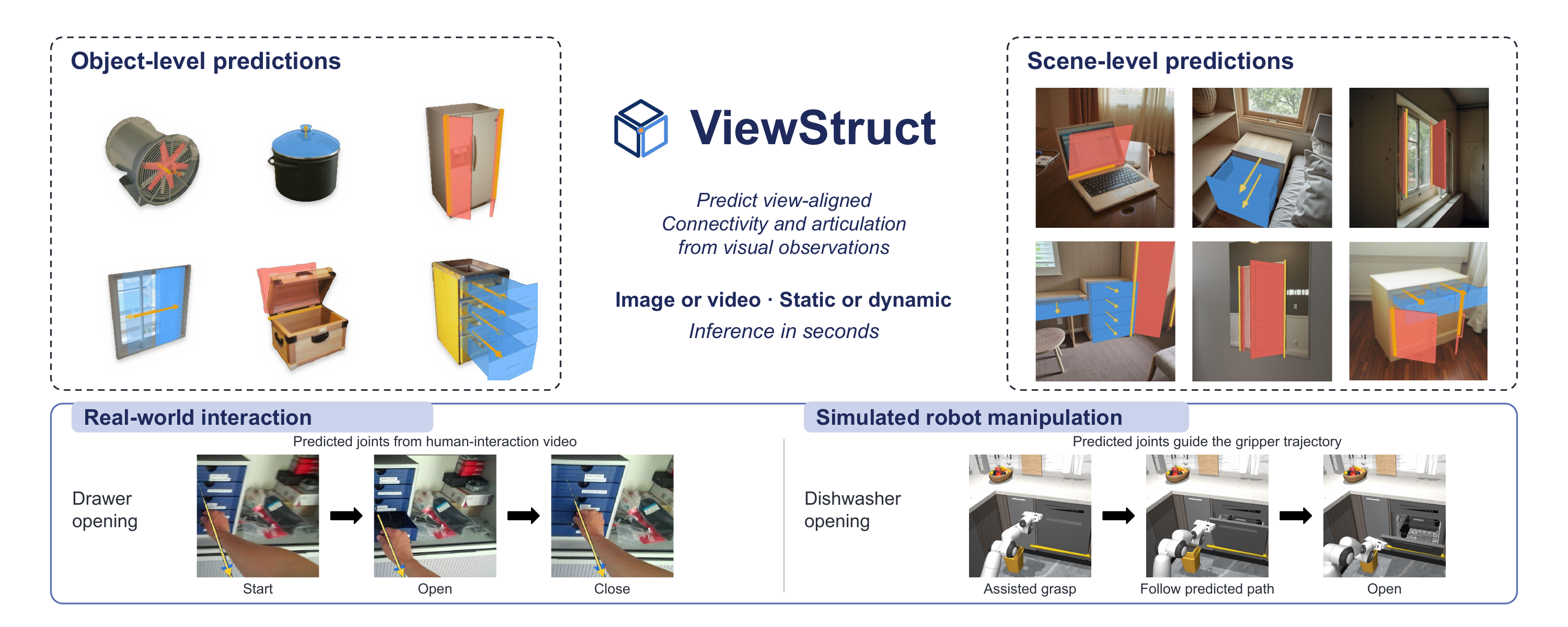}
    \caption{\textbf{ViewStruct predicts part connectivity and articulation from visual observations.} Given a reference RGB image with aligned part segmentation and depth, optionally supplemented by images or video, it predicts structure in seconds. Part identifiers associate predictions with observed regions, while joint geometry is expressed in a gravity-aligned, view-relative frame. Examples show object- and scene-level predictions, transfer to real-world interactions, and predicted trajectories used for simulated manipulation.}
    \label{fig:teaser}
\end{figure}
% !TEX root = main.tex

% \begin{abstract}
% Inferring how object parts connect and move is essential for building interactive scene representations and planning robotic manipulation. We present ViewStruct, a view-grounded framework for jointly predicting part connectivity and articulation in objects and scenes. Given a reference RGB image with aligned part segmentation and depth, optionally supplemented by additional images or video, ViewStruct predicts rigid-body connectivity and joint parameters in a gravity-aligned, view-relative frame. The framework combines visual and semantic priors for direct autoregressive prediction of articulated structure in seconds without per-instance optimisation. Observed articulation motion is optional, and additional motion evidence is incorporated when available. Experiments demonstrate improved connectivity and articulation prediction over existing baselines, with generalisation to held-out object categories and unseen asset sources. Despite training exclusively on synthetic data, ViewStruct also transfers to real-world human-interaction videos. Finally, simulated manipulation experiments show that the predicted joints provide useful motion priors for robotic interaction.

% \end{abstract}
\begin{abstract}
Interacting with unfamiliar objects requires understanding how their parts connect and move. We introduce ViewStruct, a framework for jointly predicting connectivity and articulation in objects and scenes. Given an RGB image with aligned part segmentation and depth, optionally supplemented by images or video, ViewStruct predicts rigid-body groupings, attachments, and joint parameters. Its view-grounded representation links predictions to observed regions and expresses joint geometry in a gravity-aligned, view-relative frame, without requiring category-specific canonical orientations. A pretrained vision--language model is adapted to predict structure in seconds, without requiring observed articulation motion or per-instance optimisation. Experiments demonstrate improved structural prediction over evaluated baselines and transfer to unseen categories, asset sources, and real-world human-interaction videos, despite task-specific training exclusively on synthetic data. Simulated manipulation experiments further demonstrate the utility of predicted joints for generating robot motion trajectories.
Project page: \url{https://bianwenjing.github.io/viewstruct/}
\end{abstract}

\section{Introduction}

Robots interacting with unfamiliar environments must understand which parts move together and which motions their connections permit. These constraints inform manipulation planning and the construction of interactive representations from visual observations. Geometry and segmentation alone do not establish them: a drawer front, handle, and box may occupy separate regions yet move as one rigid body, while adjacent cabinet doors may move independently. In scenes, this reasoning must also distinguish mechanical attachments from proximity between unrelated objects.

Practical structural understanding requires flexible inputs and efficient inference. A single image provides appearance and spatial cues but may leave motion constraints ambiguous. Additional views reveal more structure, while video provides evidence of shared and relative motion. A useful predictor should accommodate these observations without requiring every movable part to be seen in action, and produce estimates quickly enough to inform subsequent interaction.

Existing methods address these settings through different routes. OPD~\citep{jiang2022opd} and OPDMulti~\citep{sun2024opdmulti} estimate openable-part motion from single-view images. SINGAPO~\citep{liu2025singapo} and PhysX-Anything~\citep{cao2025physx} generate articulated assets from images, while Particulate~\citep{li2026particulate} infers articulation from geometry. Interaction-based methods such as ArtiPoint~\citep{werby2025articulated} and MoMa-SG~\citep{buechner2026momasg} recover articulation from observed motion. These methods typically target a particular observation setting, such as single-image prediction or estimation from observed motion. We instead use one predictor and output representation across static images and interaction sequences, allowing additional observations to refine the estimate without making motion a prerequisite.

We introduce \textbf{ViewStruct}, a VLM-based framework taking a reference RGB image with aligned part segmentation and depth, optionally supplemented by images or video. The reference observation determines the part identities and output coordinate frame. Input regions need not correspond to rigid bodies: the model predicts which regions move together and how these bodies attach. By adapting a pretrained VLM, ViewStruct combines knowledge of part functions and plausible mechanical relationships with observation-specific evidence, predicting structure in seconds without per-instance optimisation or an external refinement loop.

Our \emph{view-grounded representation} associates predictions with observed regions through persistent part identifiers. Joint geometry is expressed in a gravity-aligned, view-relative frame, avoiding category-specific canonical orientations while stabilising the vertical direction across viewpoints. Revolute pivots are decoded from image locations and relative depth, providing explicit image correspondence. This representation applies consistently to individual objects and multi-object scenes.

Experiments demonstrate improved connectivity and articulation prediction over evaluated baselines, including transfer to held-out categories and unseen asset sources. Despite task-specific training exclusively on synthetic data, ViewStruct also transfers to real-world human-interaction videos, where motion evidence improves joint classification and localisation. We further use predicted joints to generate manipulation trajectories in simulation, testing them against reference object mechanisms under assisted contact.

We summarise our contributions as follows: \textbf{First}, we formulate joint connectivity and articulation inference using a shared view-grounded representation for objects and scenes. \textbf{Second}, we develop ViewStruct, supporting single-image, multi-image, and video inputs with inference in seconds and no requirement for observed articulation motion. \textbf{Third}, we demonstrate improved structural prediction, generalisation across categories and observation domains, and utility for robotic manipulation.

\section{Related Work}

\paragraph{Articulation inference and generation.}
Prior work estimates articulation from static observations~\citep{hu2017learning,zeng2021visual,li2020category}, including single-view openable-part detection in objects and scenes~\citep{jiang2022opd,sun2024opdmulti}. Other methods construct articulated assets from graphs~\citep{liu2024cage}, images~\citep{chen2024urdformer,liu2025singapo,cao2025physx,cao2026physx}, or meshes~\citep{li2026particulate,li2026instructparticulate}, or infer articulation through synthesised motion~\citep{lee2026articulated}. ViewStruct jointly predicts mechanical connectivity and articulation over supplied regions, using a shared view-grounded representation for objects and scenes without category-specific canonical orientations.

\paragraph{Articulation from motion and interaction.}
Motion-based approaches recover articulation from multiple observed states~\citep{jiang2022ditto,jiayi2023paris,weng2024neural}, including through active interaction in scenes~\citep{hsu2023ditto}. ArtiPoint~\citep{werby2025articulated} estimates articulation through tracking and optimisation, while Pandora~\citep{yu2026pandora} and MoMa-SG~\citep{buechner2026momasg} incorporate interaction-derived articulation into scene graphs. ViewStruct operates from static observations, incorporating additional images or motion evidence when available without per-instance fitting.

\paragraph{Language models for structured 3D prediction.}
Language models support 3D modelling through structured outputs~\citep{daimeshcoder,xu2024cad,raji2026proc3d} and agentic generation with iterative feedback~\citep{yamada2025l3go,huang2024blenderalchemy,le2024articulate,zhou2026articraft,lin2026procedura}. For articulation, Real2Code~\citep{mandi2024real2code} and URDF-Anything~\citep{li2025urdf} condition models on reconstructed part bounding boxes and point clouds, respectively. ViewStruct instead conditions a compact VLM on RGB, segmentation, and depth observations, optionally supplemented by images or video, to predict view-grounded structure in seconds without an external refinement loop.
\begin{figure}[t]
    \centering
    \includegraphics[width=\textwidth]{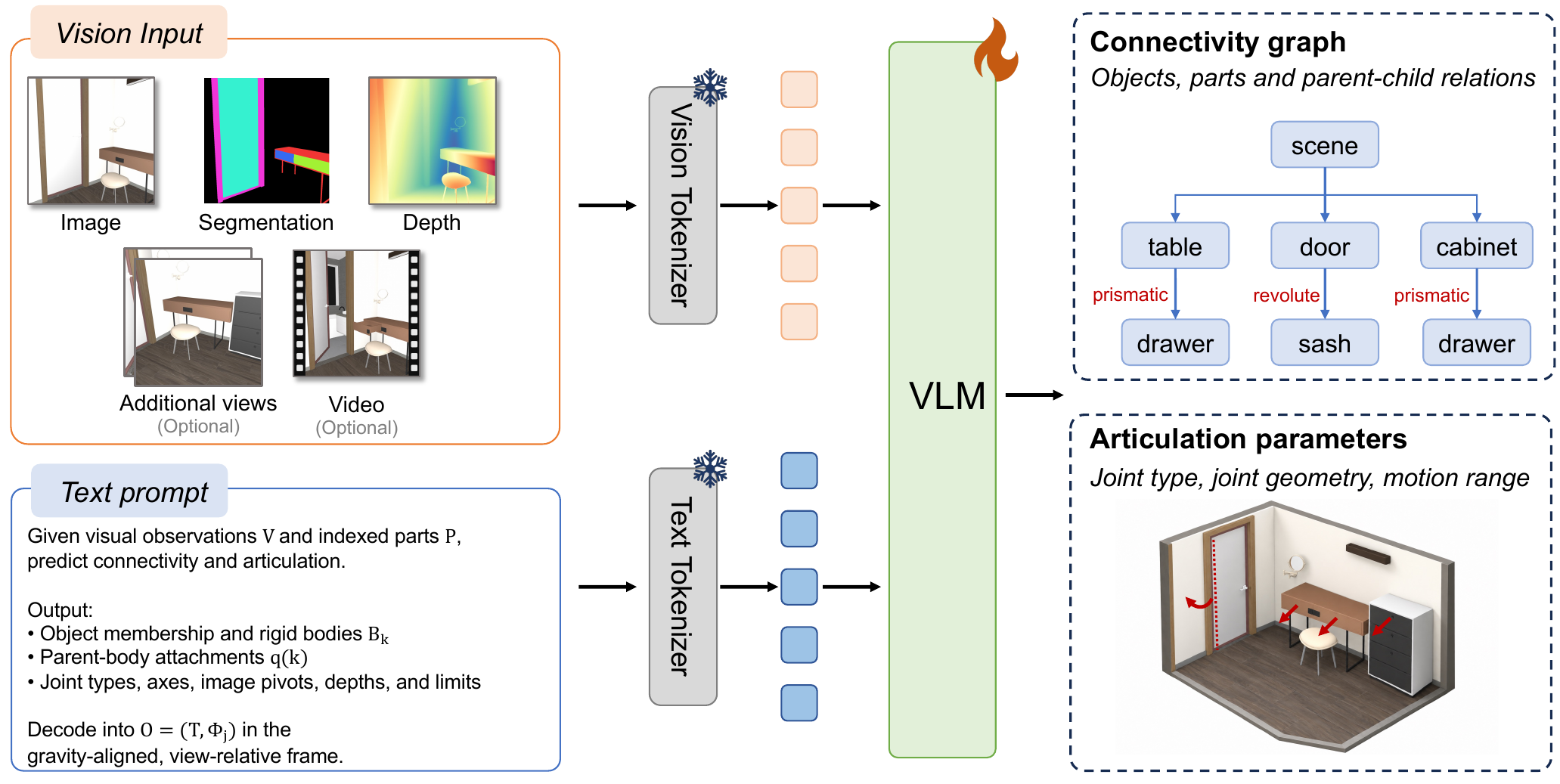}
    \caption{\textbf{Overview of ViewStruct.} Given a reference RGB image, aligned segmentation and depth maps, and a text prompt specifying an indexed part inventory, ViewStruct uses a VLM to predict part connectivity and articulation parameters in the reference-view frame. Additional views or video can provide complementary evidence, but observed articulation motion is not required.}
    \label{fig:framework}
\end{figure}
% !TEX root = main.tex
\section{Method}
\label{sec:method}

ViewStruct jointly predicts part connectivity and articulation from visual observations, using a shared representation for objects and scenes. Its inputs comprise a reference RGB image, aligned part segmentation, depth, and optional additional images or video. \Cref{fig:framework} illustrates the pipeline.

\subsection{Overall Pipeline and Problem Formulation}
\label{sec:setting}
\label{sec:input}

\paragraph{Observation and part encoding.}
We denote the visual input by $\mathcal{V}=(I,S,D,\mathcal{A})$, where $I$, $S$, and $D$ are the reference RGB image, part-segmentation map, and depth map, respectively, and $\mathcal{A}$ contains optional additional observations. Depth may be rendered, measured, or estimated. Each supplied part has a persistent identifier in $\mathcal{P}=\{0,\ldots,N-1\}$. A text prompt $Q$ specifies the task, output schema, and an indexed region inventory associating each identifier with its segmentation colour, bounding box, and relative depth interval. Additional observations provide complementary evidence while retaining the reference part identities and output frame.

\paragraph{Structural prediction.}
We adapt Qwen3-VL-4B~\citep{Qwen3-VL} to generate
\begin{equation}
    \mathcal{O}=f_{\vartheta}(\mathcal{V},Q)
    =(T,\Phi_{\!J}),
    \label{eq:repr}
\end{equation}
where $T:\mathcal{P}\rightarrow\mathcal{P}\cup\{\bot\}$ defines a forest of parent attachments and $\Phi_{\!J}$ specifies joints on articulated attachments. Attachments without joints are rigid, allowing multiple regions to form one rigid body. ViewStruct produces this structure through autoregressive decoding, without per-instance optimisation or an external refinement loop.

\subsection{View-Grounded Structure Prediction}
\label{sec:repr}
\label{sec:camnative}
\label{sec:scene}

We define the representation of the connectivity $T$ and joint parameters $\Phi_{\!J}$ in \Cref{eq:repr}. Our design addresses three ambiguities: canonical coordinates require object-specific orientation conventions, part-level parent maps impose arbitrary hierarchies within rigid bodies, and direct 3D pivot regression lacks an explicit image-space correspondence. We therefore express $\Phi_{\!J}$ in a gravity-aligned, view-relative frame, recover $T$ from joint-centred rigid groups, and reconstruct pivots from predicted image locations and depth.

\paragraph{Gravity-aligned, view-relative coordinates.}
Canonical orientations can be ambiguous for unfamiliar or symmetric objects and differ across objects in a scene. A view-relative frame provides a shared reference without requiring a category-specific front direction. Gravity alignment further reduces variation due to camera pitch and roll, simplifying the representation of vertical hinges and horizontal sliding motions.

Let $m$ be the union of the reference object's region silhouettes, $\bar{u}$ its bounding-box centre, and $\bar{d}$ its median depth. We define the frame centre and scale as
\begin{equation}
    c=\pi^{-1}(\bar{u},\bar{d}),
    \qquad
    s=\max\big(w(m),h(m)\big),
    \label{eq:frame_cs}
\end{equation}
where $\pi^{-1}$ denotes camera back-projection and $w(m),h(m)$ are metric silhouette extents at depth $\bar{d}$. Given gravity $g$ and viewing direction $v$ in camera coordinates, the frame axes are
\begin{equation}
    \hat{z}=-\frac{g}{\lVert g\rVert},
    \qquad
    \hat{x}=
    \frac{v-(v^{\!\top}\hat{z})\hat{z}}
    {\lVert v-(v^{\!\top}\hat{z})\hat{z}\rVert},
    \qquad
    \hat{y}=\hat{z}\times\hat{x}.
    \label{eq:frame_basis}
\end{equation}
With $R=[\hat{x}\;\hat{y}\;\hat{z}]^{\!\top}$, points transform as $\phi(p)=R(p-c)/s$, while directions undergo rotation only. Gravity is supplied by the renderer for synthetic observations and by the capture device's inertially aligned pose for real recordings; no external heading is required. Replacing this frame with raw camera coordinates reduces articulation success by $3.9$ percentage points (\Cref{sec:supp_ablation}). For scenes, a designated reference object determines $c$ and $s$, and all objects share the frame.

\paragraph{Joint-centred connectivity.}
Visual regions need not correspond to distinct rigid bodies: a drawer's front, handle, and box may be segmented separately but move together. Predicting a parent for every region requires choosing an internal hierarchy among these rigidly attached parts, although different choices describe the same physical structure. We therefore predict rigid-body groups and their articulated connections, using a deterministic convention to recover the part-level parent map $T$.

Each object consists of a static base $B_0$ and one moving body $B_k$ per articulated joint. The sets $B_k\subseteq\mathcal{P}$ partition the object's part inventory, with all regions in a body moving rigidly together. Entry $k$ specifies its moving regions $B_k$, parent-body index $q(k)$, and joint parameters
\begin{equation}
    \Phi_{\!J}=\{j_k\}_{k=1}^{K},
    \qquad
    j_k=(\tau_k,\hat{a}_k,x_k,[\theta_{k,-},\theta_{k,+}]),
    \label{eq:joint}
\end{equation}
where $\tau_k$ is the joint type, $\hat{a}_k$ its unit axis, $x_k$ its revolute pivot, and $[\theta_{k,-},\theta_{k,+}]$ its motion limits. Prismatic joints omit the pivot and express displacements in units of $s$; revolute limits are in radians. A moving body may attach to the static base or another moving body, supporting nested articulation such as a door mounted within a sliding assembly.

To recover $T$, each body has an anchor region $b_k$, listed first; for moving bodies, this is the joint-carrying region. The static anchor $b_0$ becomes the root. Each moving anchor $b_k$ attaches to its parent-body anchor $b_{q(k)}$ through joint $j_k$, while the remaining regions in $B_k$ attach rigidly to $b_k$. Thus, the model predicts physically meaningful grouping and attachment relationships, and internal rigid hierarchies are assigned deterministically. Joint entries follow a fixed image-position ordering, and object membership distinguishes separate hierarchies in scenes.

Grouping also reduces output length by representing each rigid body collectively. On PhysX-Mobility, the $95$th-percentile target length decreases from $2073$ to $1329$ tokens, reducing the risk of truncation for objects with many parts.

\paragraph{Image-grounded pivots.}
To associate joint geometry explicitly with the reference image, we parameterise each revolute pivot $x_k$ in \Cref{eq:joint} by a normalised image location $u_k\in[0,1000]^2$ and a relative camera-depth offset $\delta_k$. This allows the model to localise the pivot using visual cues, such as a hinged edge, and infer its depth separately.

For an image of size $W\times H$, let $\tilde{u}_k=(Wu_{k,x}/1000,Hu_{k,y}/1000)$ denote the corresponding pixel coordinates. We decode
\begin{equation}
    z_k=c_z+\delta_k s,
    \qquad
    x_k=\phi\!\left(\pi^{-1}(\tilde{u}_k,z_k)\right),
    \label{eq:pivot_px}
\end{equation}
where $c_z$ is the camera-depth component of the frame centre. Thus, the pivot entering $\Phi_{\!J}$ is reconstructed from image-grounded predictions rather than independently regressed, ensuring projection consistency by construction. Axis directions $\hat{a}_k$ are predicted directly in the reference frame, since their image projections may be degenerate or poorly resolved. The output $\mathcal{O}=(T,\Phi_{\!J})$ therefore combines deterministically expanded connectivity with decoded joint geometry.

\subsection{Data Construction and Training}
\label{sec:training}
\label{sec:data}
\label{sec:association}
\label{sec:applications}

\paragraph{Unified supervision.}
We convert annotated articulated assets from different sources into a common set of observation-grounded input--target pairs. Fixed attachments are grouped into rigid bodies, and scene examples additionally record object membership. We render observations with persistent part identifiers linking image regions to the annotated geometry, then transform joint parameters into the reference frame with consistent units. This reconciles differences in coordinate, scale, and hierarchy conventions without requiring category-specific canonical alignment.

\paragraph{Static and dynamic observations.}
From these assets, we construct both static observations and articulation clips. Static examples supervise inference from appearance and spatial relationships, while clips provide direct evidence of relative motion. Both use the same structural targets expressed in their reference frame. We vary camera viewpoints to expose the model to different appearances and visibility patterns of the same underlying structure.

\paragraph{Part-decomposition augmentation.}
To improve robustness to different test-time segmenters, we also vary the supplied part decomposition. We either split an articulated part by a plane through its centroid into two rigidly connected regions sharing the original joint, or merge two sibling regions with compatible articulation. We then update the part inventory and re-render the segmentation map $S$, preserving the joint parameters and underlying kinematics. This exposes the model to different visual decompositions of the same mechanical structure, reducing dependence on a particular segmentation. Ablations demonstrate improved prediction performance (\Cref{sec:supp_ablation}). Corpus construction and augmentation details are provided in \Cref{sec:supp_view_corpus,sec:supp_view_supervision}.
% Table reorganization: 2026-09-20. Source audit: table_sources_20260920.json.
\section{Experiments}
\label{sec:exp}

\paragraph{Implementation details.}
We fine-tune Qwen3-VL-4B-Instruct~\cite{Qwen3-VL} with LoRA for 2 epochs on the structured prediction objective with input context length $8192$ tokens, an image pixel budget of $256\times256$,
effective batch size~$32$, and a cosine learning-rate schedule peaking at $10^{-4}$. The pretrained vision encoder and the base decoder weights are kept frozen: we train rank-$64$ LoRA adapters on the decoder together with the vision--language merger. All experiments use four NVIDIA A100 GPUs. Please refer to the supplementary for more details.

\paragraph{Metrics.}
We report \textbf{parse success} (the fraction of generations that yield a parsable structured JSON) and, on the parsed subset, \textbf{parent accuracy} and \textbf{edge $F_1$} for connectivity, \textbf{joint-presence $F_1$}, \textbf{joint-type accuracy}, \textbf{joint-axis accuracy}, \textbf{joint position error} and \textbf{joint range~IoU} for articulation. Following~\citet{le2024articulate}, each failed joint is attributed to its first violated gate (type$\,{\to}\,$axis$\,{\to}\,$origin$\,{\to}\,$limit) and the \textbf{success} rate is the fraction of matched joints that pass all four. 
% since the limit gate is loose by construction, range~IoU rather than the gate carries the information about motion limits. 
We further report \textbf{inference time} in wall-clock seconds per object on a single A100. Definitions, thresholds and normalisation conventions are given in \Cref{sec:supp_metrics}.

\paragraph{Baselines.}
We compare with the zero-shot Qwen3-VL backbone, PhysX-Anything~\citep{cao2025physx}, PhysX-Omni~\citep{cao2026physx}, URDF-Anything+~\citep{li2025urdf}, and Particulate~\citep{li2026particulate}. Their input modalities are given in the caption of \Cref{tab:physx_anything}; mesh-based methods receive reference geometry. Particulate, which predicts from geometry alone, is compared in \Cref{tab:physx_detailed}. The zero-shot backbone uses the same inputs and prompt as ViewStruct. OPDMulti~\citep{sun2024opdmulti} detects openable parts and their motion parameters in a single RGB image, which is the closest published formulation to our scene setting; we evaluate it on the room views and on real video. SINGAPO~\citep{liu2025singapo} is evaluated on its supported category subset in the supplementary, and ArtiPoint~\citep{werby2025articulated} is compared on real video. Baseline conversion, part correspondence, and timing protocols are detailed in \Cref{sec:supp_exp_protocols}. 

\begin{figure}[t]
    \centering
    \includegraphics[width=\linewidth]{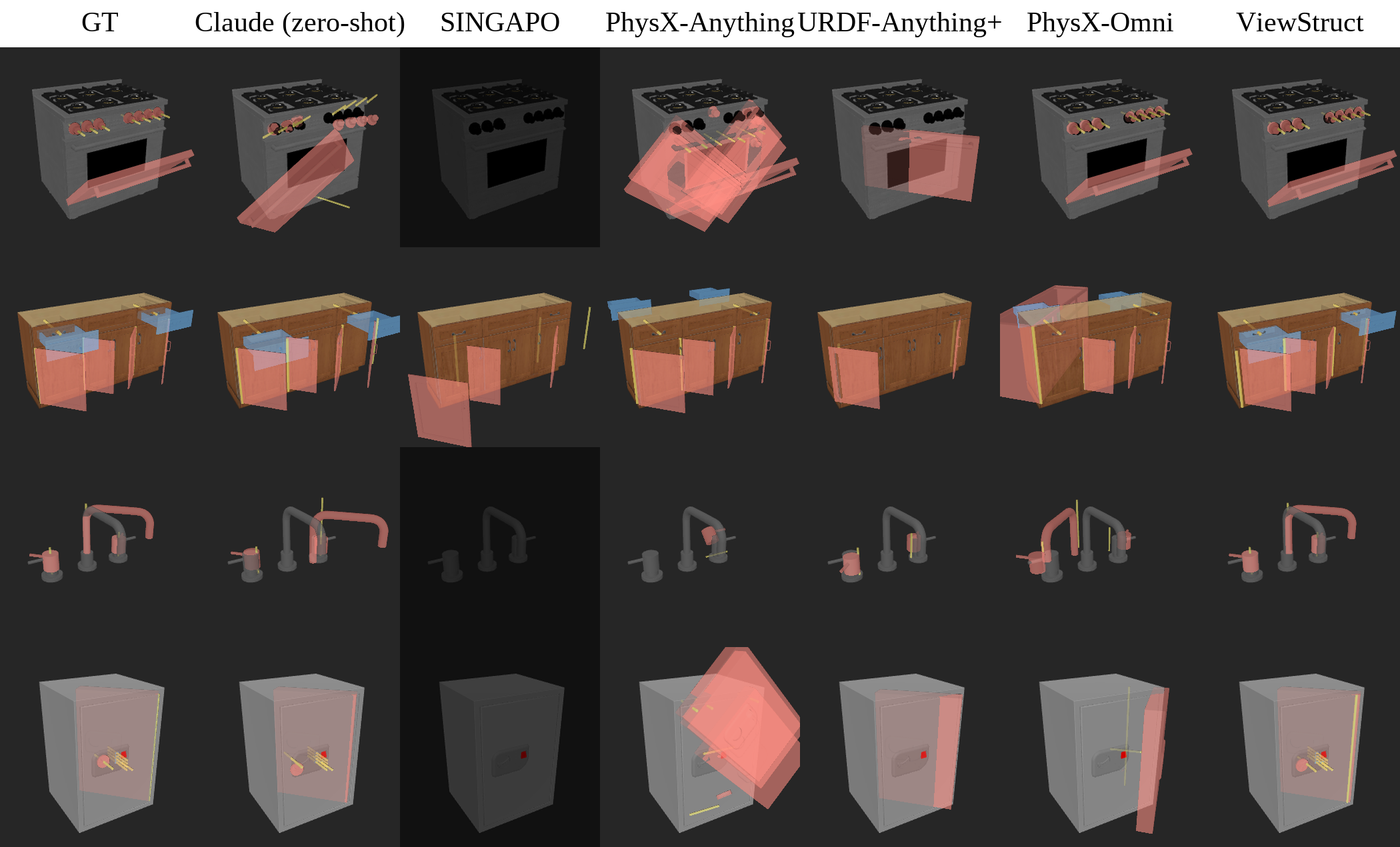}
    \caption{\textbf{Qualitative comparison on PhysX-Mobility.} Predicted motion is overlaid on the input geometry: hinges in red, slides in blue, and axes in yellow. Dimmed panels indicate missing outputs.}
    \label{fig:comparison}
\end{figure}

\subsection{Evaluation of Connectivity and Articulation Prediction}
\label{sec:physx_anything}

% We evaluate structure estimation conditioned on a single view at both the object and scene levels.

\paragraph{Object-level prediction.}
We evaluate all methods on $387$ objects from the PhysX-Mobility~\citep{cao2025physx} test split, using ViewStruct with either ground-truth depth or monocular depth estimated by MoGe~\citep{wang2025moge}. As shown in \Cref{tab:physx_anything}, ViewStruct achieves the highest joint $F_1$, while also predicting part connectivity. PhysX-Omni achieves higher type and axis accuracy and matched-joint success ($85.3\%$ versus $81.5\%$). However, these conditional metrics exclude unrecovered joints and should therefore be interpreted alongside coverage. Canonical-coordinate prediction also benefits from axis-alignment priors in this dataset, whereas ViewStruct expresses joint directions relative to the observation. When evaluated over all reference joints, ViewStruct achieves $78.2\%$ success, compared with $66.6\%$ for PhysX-Omni, with the advantage largely retained using estimated depth. Inference takes $6.5$\,s per object, compared with $241$--$267$\,s for the VLM-based joint prediction stages of PhysX-Omni and PhysX-Anything, excluding their geometry-generation time.

\Cref{fig:comparison} visualises the motions induced by the predicted joints when applied to ground-truth geometry. This highlights the effect of joint prediction errors, where inaccurate axes or pivots can cause articulated parts to tilt, detach, or follow incorrect trajectories. We additionally include Claude Fable as a qualitative baseline in \Cref{fig:comparison,fig:qual_generalisation}; due to the high inference cost of evaluating the full test set, we restrict this comparison to qualitative examples.
More detailed results, including a category-restricted comparison with SINGAPO~\cite{liu2025singapo}, are provided in \Cref{tab:physx_detailed}.
% We further report controlled ablations in \Cref{sec:supp_ablation}, where we retrain the model on this split while changing one component at a time, including the observation frame and decomposition augmentation.

\begin{table}[t]
\centering
\caption{\textbf{Structure prediction on PhysX-Mobility.} All methods receive a single reference image. ViewStruct additionally uses part segmentation and depth, either supplied or estimated by MoGe in the last row; zero-shot Qwen3-VL receives the same inputs. PhysX-Anything and PhysX-Omni use only the image, while URDF-Anything+ also receives a whole-object mesh. Comparisons with SINGAPO and Particulate appear in \Cref{tab:physx_detailed}. Success is measured over matched joints, over all reference joints (missed joints counted as failures), and per object (the fraction of objects whose reference joints are all recovered correctly; an object with many keys or buttons counts once). Pivot errors are in normalised asset units, inference times in seconds, and all other metrics in percentages. Evaluation protocols are detailed in \Cref{sec:supp_exp_protocols}.}
\label{tab:physx_anything}
\small
\setlength{\tabcolsep}{2.4pt}
\begin{tabular}{@{}lrrrrrrrrrrr@{}}
\toprule
& \multicolumn{2}{c}{\textit{Connectivity}} & \multicolumn{5}{c}{\textit{Articulation}} & \multicolumn{3}{c}{Success$\uparrow$} & \\
\cmidrule(lr){2-3} \cmidrule(lr){4-8} \cmidrule(lr){9-11}
Method & Parent$\uparrow$ & Edge $F_1\uparrow$ & Joint $F_1\uparrow$ & Type$\uparrow$ & Axis$\uparrow$ & Pivot$\downarrow$ & Range$\uparrow$ & matched & all joints & objects & Time\,(s)$\downarrow$ \\
\midrule
Qwen3-VL-4B (zero-shot) & 19.2 & 19.9 & 31.5 & 66.0 & 41.2 & 0.391 & 30.5 & 1.1 & 0.2 & 1.0 & -- \\
PhysX-Anything & 61.0 & 67.9 & 85.9 & 96.2 & 94.4 & 0.118 & 26.3 & 81.6 & 64.3 & 40.2 & 267 \\
PhysX-Omni & 70.1 & 73.6 & 85.9 & \textbf{97.8} & \textbf{95.8} & 0.096 & 26.5 & \textbf{85.3} & 66.6 & 38.9 & 241 \\
URDF-Anything+ & 17.3 & 14.7 & 18.6 & 73.0 & 58.5 & 0.269 & 34.1 & 14.0 & 1.5 & 2.3 & 64 \\
ViewStruct & \textbf{92.3} & \textbf{92.7} & \textbf{96.8} & 96.6 & 89.9 & \textbf{0.084} & \textbf{72.3} & 81.5 & \textbf{78.2} & \textbf{52.5} & \textbf{6.5} \\
ViewStruct (MoGe depth) & \textbf{92.3} & 91.0 & 96.5 & 95.6 & 89.3 & 0.094 & 69.1 & 78.3 & 74.8 & 48.7 & \textbf{6.5} \\
\bottomrule
\end{tabular}
\end{table}

\paragraph{Scene-level prediction.}\label{sec:scenes}

\begin{figure}[t]
\centering
\begin{minipage}[t]{0.48\linewidth}
\centering
    \includegraphics[width=\linewidth]{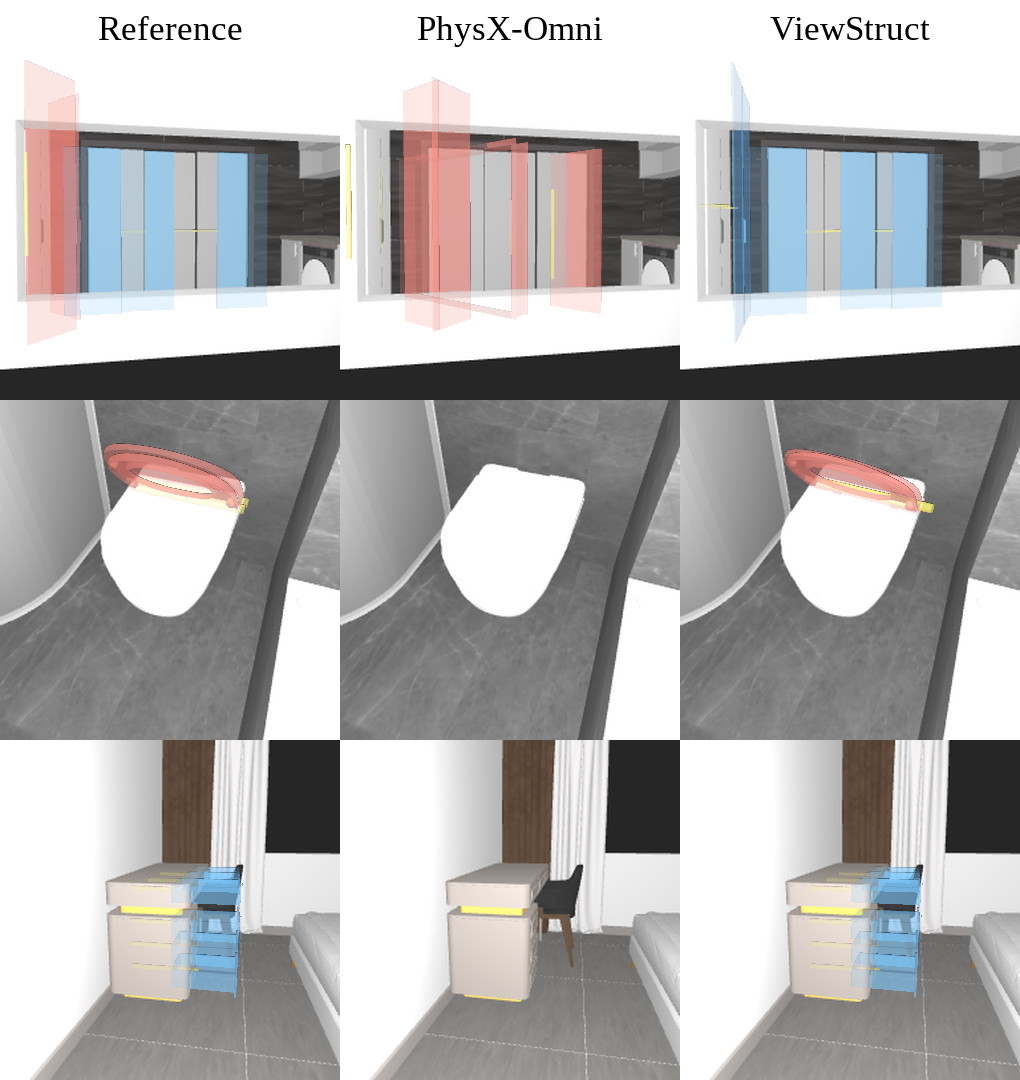}
    \caption{\textbf{GRScenes scene predictions.} Reference, object-wise PhysX-Omni, and ViewStruct results. Red and blue indicate hinged and sliding motion; yellow marks joint axes.}
    \label{fig:scene_qual}
\end{minipage}\hfill
\begin{minipage}[t]{0.48\linewidth}
\centering
    \includegraphics[width=\linewidth]{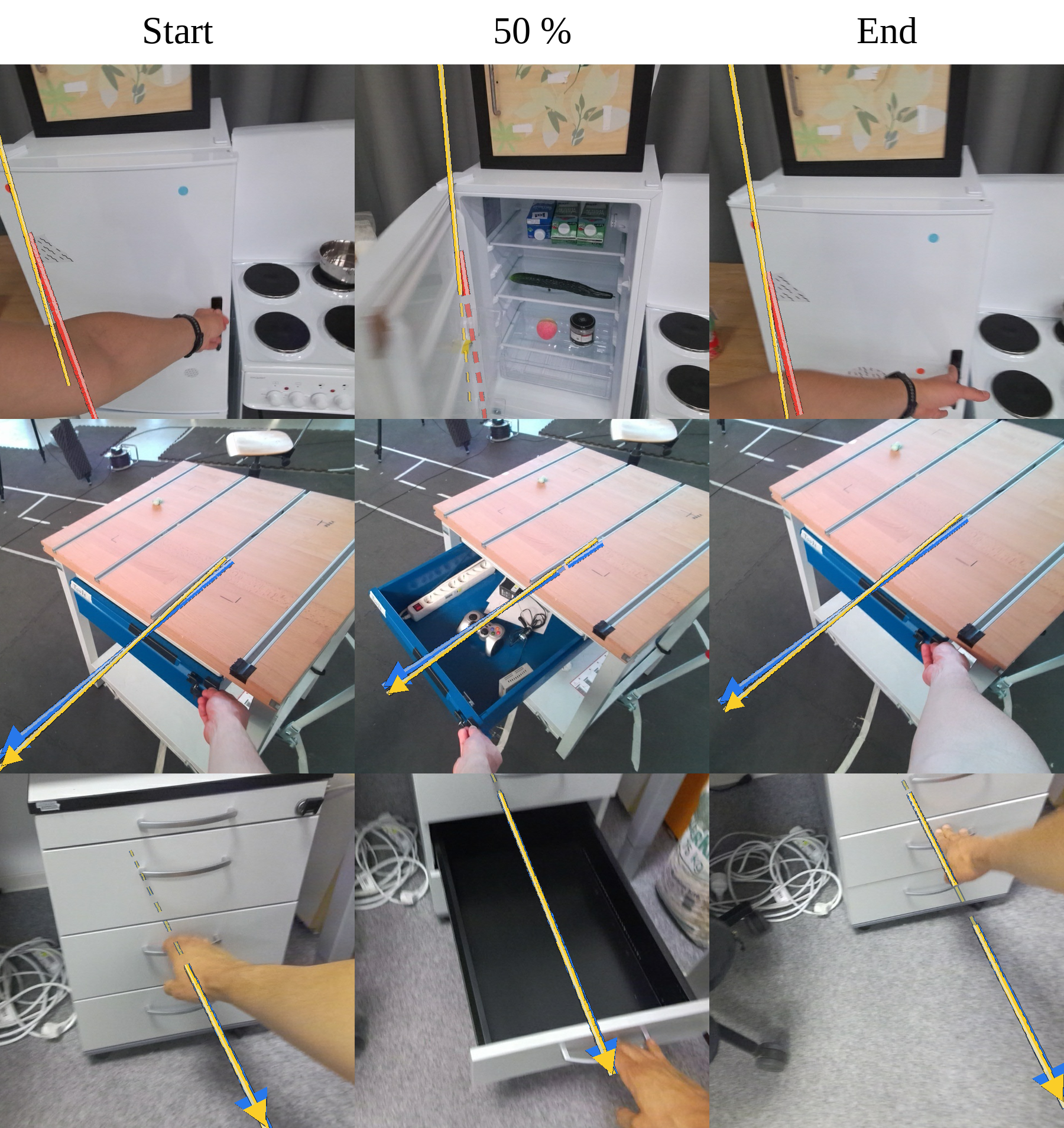}
    \caption{\textbf{Transfer to real Arti4D interactions.} Reference axes (red: hinge; blue: slide) and predicted axes (yellow) across three frames. Dashed segments are occluded.}
    \label{fig:qual_arti4d}
\end{minipage}
\end{figure}

We evaluate $559$ GRScenes~\cite{grutopia} room views containing multiple articulated objects. We compare against OPDMulti~\citep{sun2024opdmulti}, which detects openable parts in multi-object images, and PhysX-Omni~\citep{cao2026physx}, the strongest baseline by articulation success in \Cref{tab:physx_anything}. We adapt PhysX-Omni to scenes by processing each object separately using a masked input image.

As shown in \Cref{tab:scenes}, ViewStruct recovers $94.7\%$ of reference joints, compared with $15.7\%$ for OPDMulti and $30.3\%$ for PhysX-Omni. It also achieves higher type and axis accuracy and lower pivot error on matched joints. Although matched-joint success remains limited at $26.2\%$, its substantially higher coverage yields $24.8\%$ success over all reference joints, compared with $1.8\%$ and $7.4\%$ for the baselines. These results highlight the importance of evaluating both joint recovery and parameter accuracy. The last row of \Cref{tab:scenes} supplies additional evidence at test time: two further renderings of the same room, one from the same viewpoint and one from a second viewpoint, with the articulated parts moved. Joint recovery is already saturated and does not change, while type and axis accuracy improve ($77.6\rightarrow80.5$ and $62.1\rightarrow64.4$) and success over all reference joints rises from $24.8\%$ to $26.6\%$. The hinge pivot is unaffected ($0.331\rightarrow0.337$), consistent with it remaining the dominant error. \Cref{fig:scene_qual} illustrates recovered structures and a joint-type error; detailed protocols and in-frame diagnostics are provided in the supplementary.

\begin{table}[t]
\centering
\caption{\textbf{Scene-level articulation prediction on GRScenes.} Pivot thresholds use per-object normalisation. Success over matched and all reference joints distinguishes parameter accuracy from coverage. Pivot error is in asset units; other entries are percentages. The last row uses the same model and reference views with two additional room renderings showing moved parts, one from the reference viewpoint and one from another; other rows use a single view. Parent and edge scores are omitted because the predominantly static attachment structure allows a prediction with no joints to score $99.9\%$ (\Cref{sec:supp_metrics}).}
\label{tab:scenes}
\small
\setlength{\tabcolsep}{5.4pt}
\begin{tabular}{@{}lrrrrrrrrr@{}}
\toprule
Method & Parse$\uparrow$ & Recall$\uparrow$ & Joint $F_1\uparrow$ & Type$\uparrow$ & Axis$\uparrow$ & Pivot$\downarrow$ & Range$\uparrow$ & Success$\uparrow$ & Success-all$\uparrow$ \\
\midrule
OPDMulti & \textbf{100.0} & 15.7 & 25.8 & 73.9 & 49.4 & 0.440 & 50.0 & 11.8 & 1.8 \\
PhysX-Omni & \textbf{100.0} & 30.3 & 40.4 & 66.7 & 49.7 & 0.400 & 8.1 & 24.4 & 7.4 \\
ViewStruct & 98.9 & 94.7 & \textbf{95.2} & 77.6 & 62.1 & \textbf{0.331} & 72.2 & 26.2 & 24.8 \\
\quad $+2$ views with parts moved & 97.7 & \textbf{94.8} & \textbf{95.2} & \textbf{80.5} & \textbf{64.4} & 0.337 & \textbf{74.1} & \textbf{28.0} & \textbf{26.6} \\
\bottomrule
\end{tabular}
\end{table}

\subsection{Generalisation to Unseen Objects and Real-World Videos}
\label{sec:generalisation}

% We assess generalisation across object categories, asset sources, and real-world observations.

\paragraph{Unseen object categories and asset sources.}
We evaluate on $242$ objects from SimReady~\citep{lightwheel2025simready}, an unseen asset source, and $291$ objects from GRScenes~\citep{grutopia} spanning seven categories held out from its training partition. Split construction and input protocols are detailed in \Cref{sec:supp_exp_protocols}. As shown in \Cref{tab:generalisation}, ViewStruct achieves stronger generalisation to unseen asset sources and object categories than the baselines, recovering more joints while accurately predicting the type, axis, pivot, and range of matched joints. Since it recovers nearly all reference joints, its two success rates almost coincide ($33.3\%$ and $31.5\%$ on SimReady, $48.7\%$ and $48.5\%$ on the held-out categories), whereas the generative baselines lose more than half of their matched-joint rate once missed joints are counted.

\begin{table}[t]
\centering
\caption{\textbf{Object-level generalisation to SimReady and held-out GRScenes categories.} Metrics, units and columns follow \Cref{tab:physx_anything}: success is measured over matched joints and over all reference joints (missed joints counted as failures).}
\label{tab:generalisation}
\normalsize
\setlength{\tabcolsep}{9.3pt}
\begin{tabular}{@{}lrrrrrrr@{}}
\toprule
Method & Joint $F_1\uparrow$ & Type$\uparrow$ & Axis$\uparrow$ & Pivot$\downarrow$ & Range$\uparrow$ & Success$\uparrow$ & Success-all$\uparrow$ \\
\midrule
\multicolumn{8}{@{}l}{\textit{SimReady objects ($242$)}} \\
Qwen3-VL-4B (zero-shot) & 17.9 & 46.7 & 39.3 & 0.382 & 24.6 & 0.0 & 0.0 \\
PhysX-Anything & 61.3 & 62.0 & 48.7 & 0.327 & 15.9 & 14.9 & 6.9 \\
URDF-Anything+ & 33.0 & 65.7 & 52.1 & 0.239 & 31.1 & 2.8 & 0.6 \\
PhysX-Omni & 61.1 & 62.8 & 52.6 & 0.380 & 18.1 & 18.0 & 8.2 \\
ViewStruct & \textbf{95.0} & \textbf{75.0} & \textbf{67.1} & \textbf{0.115} & \textbf{31.5} & \textbf{33.3} & \textbf{31.5} \\
\midrule
\multicolumn{8}{@{}l}{\textit{GRScenes objects: held-out categories ($291$)}} \\
Qwen3-VL-4B (zero-shot) & 43.4 & 33.7 & 25.6 & 0.229 & 22.6 & 0.0 & 0.0 \\
PhysX-Anything & 52.5 & 61.1 & 29.9 & 0.432 & 13.8 & 8.9 & 3.3 \\
URDF-Anything+ & 26.6 & 37.6 & 25.6 & 0.386 & 28.7 & 0.8 & 0.1 \\
PhysX-Omni & 48.0 & 54.8 & 21.4 & 0.494 & 5.7 & 16.0 & 5.2 \\
ViewStruct & \textbf{99.7} & \textbf{71.8} & \textbf{48.7} & \textbf{0.226} & \textbf{56.9} & \textbf{48.7} & \textbf{48.5} \\
\bottomrule
\end{tabular}
\end{table}

\paragraph{Real-world interaction videos.}\label{sec:real}

We evaluate ViewStruct on $342$ human interactions from Arti4D~\citep{werby2025articulated} without additional training. The model receives an initial RGB-D observation with regions extracted by Mobile-SAM, either alone, supplemented with two interaction frames, or supplemented with both these frames and a $16$-frame interaction clip, as specified in \Cref{tab:arti4d}. We include ArtiPoint~\citep{werby2025articulated} as a reference, noting that it requires the full RGB-D sequence, camera poses, and reference interaction cues, and OPDMulti, which reads a single RGB frame and is the only baseline trained on real indoor captures. From the static observation alone, ViewStruct correctly identifies the joint type in $69.3\%$ of prismatic and $77.2\%$ of revolute interactions. Adding the clip improves these accuracies by $16.7$ and $15.8$ percentage points, respectively, and reduces hinge-position error from $0.31$ to $0.23$\,m. Under the same single-observation condition, ViewStruct exceeds OPDMulti on every column of \Cref{tab:arti4d}, by $17.1$ and $5.2$ points of type accuracy and by $0.11$\,m of hinge position. Together with the qualitative results in \Cref{fig:qual_arti4d}, these findings demonstrate transfer to real-world interactions despite task-specific training exclusively on synthetic data, without egocentric human-interaction examples during fine-tuning.

\begin{table}[t]
\centering
\caption{\textbf{Evaluation on Arti4D real-world videos.} All ViewStruct variants receive an initial RGB-D observation with Mobile-SAM regions; additional RGB frames and clips provide motion evidence without camera poses. ArtiPoint uses the full RGB-D sequence, camera poses, and reference interaction cues; OPDMulti uses the initial RGB image alone and returns no detection for $35$ interactions. Metrics follow~\citet{werby2025articulated}: type accuracy (\%) for each reference joint type, undirected axis error in degrees, and hinge-position error as the common-perpendicular distance in metres. Errors are reported as means, with medians in parentheses; $n$ denotes the number of evaluated outputs.}
\label{tab:arti4d}
\small
\setlength{\tabcolsep}{8pt}
\begin{tabular}{@{}llcccccc@{}}
\toprule
& & \multicolumn{2}{c}{Type acc.\,$\uparrow$} & \multicolumn{2}{c}{Axis error\,$\downarrow$} & Hinge pos.\,$\downarrow$ & $n$ \\
\cmidrule(lr){3-4} \cmidrule(lr){5-6}
Method & Visual input & prism. & revol. & prism. & revol. & (m) & \\
\midrule
ArtiPoint & Full RGB-D sequence & 75.3 & \textbf{97.3} & \textbf{20.5$^\circ$} (\textbf{4.4$^\circ$}) & \textbf{20.9$^\circ$} (10.3$^\circ$) & \textbf{0.19} (\textbf{0.12}) & 338 \\
OPDMulti & Initial RGB only & 52.2 & 72.0 & 44.3$^\circ$ (32.5$^\circ$) & 42.8$^\circ$ (30.9$^\circ$) & 0.42 (0.37) & 307 \\
ViewStruct & Initial RGB-D only & 69.3 & 77.2 & 36.7$^\circ$ (17.0$^\circ$) & 37.6$^\circ$ (4.0$^\circ$) & 0.31 (0.22) & 342 \\
ViewStruct & Initial RGB-D + 2 frames & 81.5 & 91.2 & 27.3$^\circ$ (14.2$^\circ$) & 27.0$^\circ$ (1.8$^\circ$) & 0.27 (0.18) & 341 \\
ViewStruct & Initial RGB-D + 2 frames + clip & \textbf{86.0} & 93.0 & 28.6$^\circ$ (15.8$^\circ$) & 21.3$^\circ$ (\textbf{1.6$^\circ$}) & 0.23 (0.18) & 342 \\
\bottomrule
\end{tabular}
\end{table}

\subsection{Downstream Manipulation}
\label{sec:manip}

ViewStruct provides camera-relative articulation estimates that can guide robot motion from a single observation, without requiring a prior interaction demonstration. Its predicted axes, pivots, and ranges define candidate opening trajectories for doors and drawers.

We evaluate these trajectories on PhysX-Mobility in simulation, retaining each object's reference mechanism while a compliant gripper follows the predicted motion. \Cref{tab:manip} reports opening success over all reference joints, counting missing predictions as failures, with an optional reverse-direction retry. Supplied contact locations and simplified physics isolate articulation quality; controller settings and success thresholds are detailed in \Cref{sec:supp_manip_protocol}.

\begin{figure}[t]
    \centering
    \includegraphics[width=\linewidth]{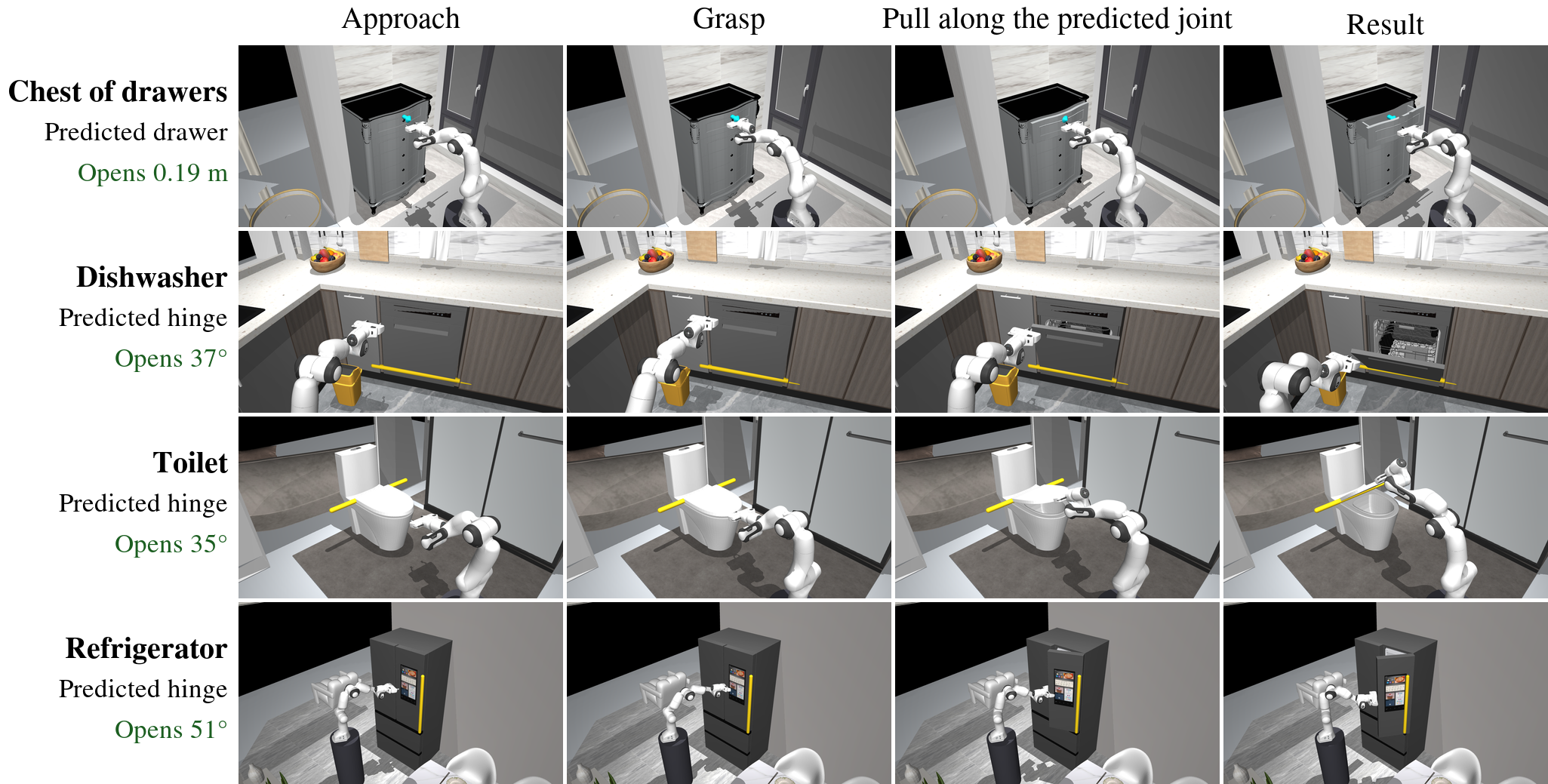}
    \caption{\textbf{Panda manipulation in GRScenes.} The arm follows ViewStruct predictions while objects retain reference mechanisms. Cyan arrows indicate sliding; yellow bars mark hinge axes.}
    \label{fig:mujoco}
    \vspace{-7pt}
\end{figure}

We further evaluate a Panda arm in MuJoCo across $10$ GRScenes rooms, again preserving the reference mechanisms. Eight of ten trials reach the opening threshold. \Cref{fig:mujoco} shows representative executions; assisted-grasp settings, additional examples are provided in \Cref{sec:supp_manip_protocol,fig:mujoco_supp}.

\begin{wraptable}{r}{0.40\linewidth}
\vspace{-\intextsep}
\centering
\small
\setlength{\tabcolsep}{6pt}
\caption{\textbf{Compliant manipulation on PhysX-Mobility.} Success (\%) over all reference joints; missing predictions count as failures. ``Attempted'' reports prediction coverage, ``$+$\,retry'' allows one reversal of the motion direction. Per-joint-type results are in \Cref{tab:manip_full}.}
\label{tab:manip}
\begin{tabular}{@{}lrrr@{}}
\toprule
Method & Attempted & Opens & +\,retry \\
\midrule
GT joints (oracle) & 100.0 & 99.9 & 99.9 \\
\midrule
PhysX-Anything & 57.2 & 50.3 & 52.3 \\
URDF-Anything+ & 11.3 & 4.4 & 4.9 \\
PhysX-Omni & 56.5 & 48.9 & 51.2 \\
ViewStruct & 91.3 & \textbf{74.6} & \textbf{80.3} \\
\bottomrule
\end{tabular}
\end{wraptable}

\section{Conclusion}

We presented ViewStruct, a view-grounded framework for jointly predicting part connectivity and articulation in objects and scenes. By combining visual and semantic priors, ViewStruct performs direct autoregressive prediction without per-instance optimisation, inferring structure from a single static observation and incorporating motion evidence when available. Experiments demonstrate improved structural prediction over existing baselines, generalisation to unseen object categories and asset sources, and transfer from synthetic training data to real-world interaction videos. Simulated manipulation experiments further show that the predicted joints can guide effective interaction trajectories. These results highlight the potential of view-grounded structural inference for connecting visual observations to actionable representations of articulated environments.

\bibliography{mybib}

@String(TOG   = {ACM Trans. Graph.})

@String(TOG   = {ACM TOG})

@inproceedings{daimeshcoder,
  title={MeshCoder: LLM-Powered Structured Mesh Code Generation from Point Clouds},
  author={Dai, BingQuan and Li, Luo and Tang, Qihong and Wang, Jie and Lian, Xinyu and Xu, Hao and Qin, Minghan and XU, Xudong and Dai, Bo and Wang, Haoqian and others},
  booktitle={The Thirty-ninth Annual Conference on Neural Information Processing Systems},
  year={2025}
}

@article{cao2025physx,
  title={PhysX-Anything: Simulation-Ready Physical 3D Assets from Single Image},
  author={Cao, Ziang and Hong, Fangzhou and Chen, Zhaoxi and Pan, Liang and Liu, Ziwei},
  journal={arXiv preprint arXiv:2511.13648},
  year={2025}
}

@article{le2024articulate,
  title={Articulate-anything: Automatic modeling of articulated objects via a vision-language foundation model},
  author={Le, Long and Xie, Jason and Liang, William and Wang, Hung-Ju and Yang, Yue and Ma, Yecheng Jason and Vedder, Kyle and Krishna, Arjun and Jayaraman, Dinesh and Eaton, Eric},
  journal={arXiv preprint arXiv:2410.13882},
  year={2024}
}

@article{li2025urdf,
  title={URDF-Anything: Constructing Articulated Objects with 3D Multimodal Language Model},
  author={Li, Zhe and Bai, Xiang and Zhang, Jieyu and Wu, Zhuangzhe and Xu, Che and Li, Ying and Hou, Chengkai and Zhang, Shanghang},
  journal={arXiv preprint arXiv:2511.00940},
  year={2025}
}

@article{hu2017learning,
  title={Learning to predict part mobility from a single static snapshot},
  author={Hu, Ruizhen and Li, Wenchao and Van Kaick, Oliver and Shamir, Ariel and Zhang, Hao and Huang, Hui},
  journal={ACM Transactions On Graphics (TOG)},
  volume={36},
  number={6},
  pages={1--13},
  year={2017},
  publisher={ACM New York, NY, USA}
}

@inproceedings{zeng2021visual,
  title={Visual identification of articulated object parts},
  author={Zeng, Vicky and Lee, Tabitha Edith and Liang, Jacky and Kroemer, Oliver},
  booktitle={2021 IEEE/RSJ International Conference on Intelligent Robots and Systems (IROS)},
  pages={2443--2450},
  year={2021},
  organization={Ieee}
}

@inproceedings{jiang2022ditto,
  title={Ditto: Building digital twins of articulated objects from interaction},
  author={Jiang, Zhenyu and Hsu, Cheng-Chun and Zhu, Yuke},
  booktitle={Proceedings of the IEEE/CVF Conference on Computer Vision and Pattern Recognition},
  pages={5616--5626},
  year={2022}
}

@article{mandi2024real2code,
  title={Real2code: Reconstruct articulated objects via code generation},
  author={Mandi, Zhao and Weng, Yijia and Bauer, Dominik and Song, Shuran},
  journal={arXiv preprint arXiv:2406.08474},
  year={2024}
}

@article{xu2024cad,
  title={Cad-mllm: Unifying multimodality-conditioned cad generation with mllm},
  author={Xu, Jingwei and Wang, Chenyu and Zhao, Zibo and Liu, Wen and Ma, Yi and Gao, Shenghua},
  journal={arXiv preprint arXiv:2411.04954},
  year={2024}
}

@article{Qwen3-VL,
      title={Qwen3-VL Technical Report}, 
      author={Shuai Bai and Yuxuan Cai and Ruizhe Chen and Keqin Chen and Xionghui Chen and Zesen Cheng and Lianghao Deng and Wei Ding and Chang Gao and Chunjiang Ge and Wenbin Ge and Zhifang Guo and Qidong Huang and Jie Huang and Fei Huang and Binyuan Hui and Shutong Jiang and Zhaohai Li and Mingsheng Li and Mei Li and Kaixin Li and Zicheng Lin and Junyang Lin and Xuejing Liu and Jiawei Liu and Chenglong Liu and Yang Liu and Dayiheng Liu and Shixuan Liu and Dunjie Lu and Ruilin Luo and Chenxu Lv and Rui Men and Lingchen Meng and Xuancheng Ren and Xingzhang Ren and Sibo Song and Yuchong Sun and Jun Tang and Jianhong Tu and Jianqiang Wan and Peng Wang and Pengfei Wang and Qiuyue Wang and Yuxuan Wang and Tianbao Xie and Yiheng Xu and Haiyang Xu and Jin Xu and Zhibo Yang and Mingkun Yang and Jianxin Yang and An Yang and Bowen Yu and Fei Zhang and Hang Zhang and Xi Zhang and Bo Zheng and Humen Zhong and Jingren Zhou and Fan Zhou and Jing Zhou and Yuanzhi Zhu and Ke Zhu},
	  journal={arXiv preprint arXiv:2511.21631},
      year={2025}
}

@article{raji2026proc3d,
  title={Proc3D: Procedural 3D Generation and Parametric Editing of 3D Shapes with Large Language Models},
  author={Raji, Fadlullah and Petrangeli, Stefano and Gadelha, Matheus and Shen, Yu and Bhattacharya, Uttaran and Wu, Gang},
  journal={arXiv preprint arXiv:2601.12234},
  year={2026}
}

@inproceedings{yamada2025l3go,
  title={L3go: Language agents with chain-of-3d-thoughts for generating unconventional objects},
  author={Yamada, Yutaro and Chandu, Khyathi and Lin, Bill Yuchen and Hessel, Jack and Yildirim, Ilker and Choi, Yejin},
  booktitle={Proceedings of the 2025 Conference of the Nations of the Americas Chapter of the Association for Computational Linguistics: Human Language Technologies (System Demonstrations)},
  pages={456--469},
  year={2025}
}

@inproceedings{huang2024blenderalchemy,
  title={Blenderalchemy: Editing 3d graphics with vision-language models},
  author={Huang, Ian and Yang, Guandao and Guibas, Leonidas},
  booktitle={European Conference on Computer Vision},
  pages={297--314},
  year={2024},
  organization={Springer}
}

@inproceedings{weng2024neural,
  title={Neural implicit representation for building digital twins of unknown articulated objects},
  author={Weng, Yijia and Wen, Bowen and Tremblay, Jonathan and Blukis, Valts and Fox, Dieter and Guibas, Leonidas and Birchfield, Stan},
  booktitle={Proceedings of the IEEE/CVF Conference on Computer Vision and Pattern Recognition},
  pages={3141--3150},
  year={2024}
}

@article{chen2024urdformer,
  title={Urdformer: A pipeline for constructing articulated simulation environments from real-world images},
  author={Chen, Zoey and Walsman, Aaron and Memmel, Marius and Mo, Kaichun and Fang, Alex and Vemuri, Karthikeya and Wu, Alan and Fox, Dieter and Gupta, Abhishek},
  journal={arXiv preprint arXiv:2405.11656},
  year={2024}
}

@inproceedings{liu2024cage,
  title={Cage: Controllable articulation generation},
  author={Liu, Jiayi and Tam, Hou In Ivan and Mahdavi-Amiri, Ali and Savva, Manolis},
  booktitle={Proceedings of the IEEE/CVF Conference on Computer Vision and Pattern Recognition},
  pages={17880--17889},
  year={2024}
}

@inproceedings{liu2025singapo,
  title={Singapo: Single image controlled generation of articulated parts in objects},
  author={Liu, Jiayi and Iliash, Denys and Chang, Angel and Savva, Manolis and Mahdavi Amiri, Ali},
  booktitle={International Conference on Learning Representations},
  volume={2025},
  pages={97511--97532},
  year={2025}
}

@inproceedings{li2026particulate,
  title={Particulate: Feed-forward 3d object articulation},
  author={Li, Ruining and Yao, Yuxin and Zheng, Chuanxia and Rupprecht, Christian and Lasenby, Joan and Wu, Shangzhe and Vedaldi, Andrea},
  booktitle={Proceedings of the IEEE/CVF Conference on Computer Vision and Pattern Recognition},
  pages={27708--27718},
  year={2026}
}

@article{werby2025articulated,
   author={Werby, Abdelrhman and Buechner, Martin and Roefer, Adrian and Huang, Chenguang and Burgard, Wolfram and Valada, Abhinav},
   title={Articulated Object Estimation in the Wild},
   journal={Conference on Robot Learning (CoRL)},
   year={2025},
}

@article{buechner2026momasg,
  title={Articulated 3D Scene Graphs for Open-World Mobile Manipulation},
  author={Buechner, Martin and Roefer, Adrian and Engelbracht, Tim
          and Welschehold, Tim and Bauer, Zuria and Blum, Hermann
          and Pollefeys, Marc and Valada, Abhinav},
  journal={arXiv preprint arXiv:2602.16356},
  year={2026}
}

@inproceedings{grutopia,
    title={GRUtopia: Dream General Robots in a City at Scale},
    author={Wang, Hanqing and Chen, Jiahe and Huang, Wensi and Ben, Qingwei and Wang, Tai and Mi, Boyu and Huang, Tao and Zhao, Siheng and Chen, Yilun and Yang, Sizhe and Cao, Peizhou and Yu, Wenye and Ye, Zichao and Li, Jialun and Long, Junfeng and Wang, ZiRui and Wang, Huiling and Zhao, Ying and Tu, Zhongying and Qiao, Yu and Lin, Dahua and Pang Jiangmiao},
    year={2024},
    booktitle={arXiv},
}

@misc{lightwheel2025simready,
  title={SimReady: Simulation-Ready 3D Assets},
  author={{Lightwheel}},
  howpublished={\url{https://simready.com/}},
  year={2025}
}

@inproceedings{wang2025moge,
  title={Moge: Unlocking accurate monocular geometry estimation for open-domain images with optimal training supervision},
  author={Wang, Ruicheng and Xu, Sicheng and Dai, Cassie and Xiang, Jianfeng and Deng, Yu and Tong, Xin and Yang, Jiaolong},
  booktitle={Proceedings of the Computer Vision and Pattern Recognition Conference},
  pages={5261--5271},
  year={2025}
}

@article{cao2026physx,
  title={PhysX-Omni: Unified Simulation-Ready Physical 3D Generation for Rigid, Deformable, and Articulated Objects},
  author={Cao, Ziang and Liu, Yinghao and Li, Haitian and Yao, Runmao and Hong, Fangzhou and Chen, Zhaoxi and Pan, Liang and Liu, Ziwei},
  journal={arXiv preprint arXiv:2605.21572},
  year={2026}
}

@inproceedings{sun2024opdmulti,
  title={Opdmulti: Openable part detection for multiple objects},
  author={Sun, Xiaohao and Jiang, Hanxiao and Savva, Manolis and Chang, Angel},
  booktitle={2024 International Conference on 3D Vision (3DV)},
  pages={169--178},
  year={2024},
  organization={IEEE}
}

@inproceedings{jiang2022opd,
  title={OPD: Single-view 3D openable part detection},
  author={Jiang, Hanxiao and Mao, Yongsen and Savva, Manolis and Chang, Angel X},
  booktitle={Computer Vision--ECCV 2022: 17th European Conference, Tel Aviv, Israel, October 23--27, 2022, Proceedings, Part XXXIX},
  pages={410--426},
  year={2022},
  organization={Springer}
}

@inproceedings{li2020category,
  title={Category-level articulated object pose estimation},
  author={Li, Xiaolong and Wang, He and Yi, Li and Guibas, Leonidas J and Abbott, A Lynn and Song, Shuran},
  booktitle={Proceedings of the IEEE/CVF conference on computer vision and pattern recognition},
  pages={3706--3715},
  year={2020}
}

@inproceedings{jiayi2023paris,
  title={PARIS: Part-level Reconstruction and Motion Analysis for Articulated Objects},
  author={Liu, Jiayi and Mahdavi-Amiri, Ali and Savva, Manolis},
  booktitle={Proceedings of the IEEE/CVF International Conference on Computer Vision},
  pages={352--363},
  year={2023}
}

@article{hsu2023ditto,
  title={Ditto in the house: Building articulation models of indoor scenes through interactive perception},
  author={Hsu, Cheng-Chun and Jiang, Zhenyu and Zhu, Yuke},
  journal={arXiv preprint arXiv:2302.01295},
  year={2023}
}

@article{yu2026pandora,
  title={Pandora: Articulated 3D scene graphs from egocentric vision},
  author={Yu, Alan and Chang, Yun and Xie, Christopher and Carlone, Luca},
  journal={arXiv preprint arXiv:2603.28732},
  year={2026}
}

@article{zhou2026articraft,
  title={Articraft: An agentic system for scalable articulated 3d asset generation},
  author={Zhou, Matt and Li, Ruining and Lyu, Xiaoyang and Song, Zhaomou and Huang, Zhening and Zheng, Chuanxia and Rupprecht, Christian and Vedaldi, Andrea and Wu, Shangzhe},
  journal={arXiv preprint arXiv:2605.15187},
  year={2026}
}

@inproceedings{lee2026articulated,
  title={Articulated Object Reconstruction from Rest-State Observation},
  author={Lee, Daeun and Lee, Jaeah and Kim, Woosung and Jung, Haebeom and Park, Jaesik},
  booktitle={European Conference on Computer Vision},
  pages={56--74},
  year={2026},
  organization={Springer}
}

@article{lin2026procedura,
  title   = {Agentic 3D Modeling with Procedural Control},
  author  = {Lin, Youtian and Yang, Yikang and Hu, Zhanpeng and
             Zhou, Mengqi and Zhang, Feihu and Cao, Xun and
             Liu, Jiaheng and Yao, Yao},
  journal = {arXiv preprint},
  year    = {2026},
}

@article{li2026instructparticulate,
  title   = {{Instruct-Particulate}: Scaling Feed-Forward 3D Object Articulation with Kinematic Control},
  author  = {Li, Ruining and Yao, Yuxin and Zhou, Matt and Zheng, Chuanxia and Rupprecht, Christian and Lasenby, Joan and Wu, Shangzhe and Vedaldi, Andrea},
  journal = {arXiv preprint arXiv:2606.14699},
  year    = {2026}
}
\bibliographystyle{arxiv}
\clearpage
\appendix
% !TEX root = main.tex
\section{Implementation Details}
\label{sec:supp_view_details}

This supplement describes the input and training configuration, evaluation protocols, additional results, and downstream use of the predicted structure.

\subsection{Observation Encoding and Output Conventions}
\label{sec:supp_view_input}
\label{sec:supp_view_repr}
\label{sec:supp_view_frame}

\paragraph{Input encoding.}
The reference input comprises RGB $I$, part segmentation $S$, and depth $D$. The prompt associates each part identifier $p$ with its segmentation colour, 2D bounding box in coordinates scaled to $[0,1000]$, and 5th/95th-percentile depth relative to the observation-frame centre $c$, normalised by scale $s$:
\begin{equation}
    \texttt{prompt}(I,S,D)\;\supseteq\;
    \bigl\{(p,\textit{colour}_p,\textit{box}_p,[d_p^5,d_p^{95}]) : p\in\mathcal{P}\bigr\}.
    \label{eq:prompt-camnative}
\end{equation}
Camera information defines the back-projection and gravity alignment of \Cref{eq:frame_cs,eq:frame_basis}. The centre is obtained from the reference object's silhouette-box centre at its median depth, and the scale from its largest metric silhouette extent. The horizontal axis follows the projected viewing direction. For the MoGe variant, the frame and depth intervals are recomputed from estimated depth; evaluation uses scale alignment.

Segmentation colours are permuted deterministically per asset, and part identifiers are randomly permuted during training, with the same permutation applied to the prompt and target. These operations reduce shortcuts based on colours or indices. Additional RGB views and clips retain the reference inventory and coordinate frame; they do not introduce independently indexed part lists.

\paragraph{Body and joint serialisation.}
The output lists static regions in \texttt{base} and moving bodies in \texttt{joints}. Each joint entry contains its rigidly connected \texttt{parts}, a parent-body reference \texttt{on}, and its articulation parameters. The first region in a body is its anchor; in a moving body, that region carries the joint. The remaining regions attach rigidly to the anchor. Each part appears in exactly one body, and joints are ordered by moving-part box centre, from top to bottom and then left to right. This grouping avoids predicting every internal rigid attachment, while retaining a deterministic convention for expanding the parent map. 

\paragraph{Pivot parameterisation.}
We represent each revolute pivot by a 2D image position and a relative depth, from which its 3D position is decoded using \Cref{eq:pivot_px}. This explicitly grounds the pivot in the reference observation. Joint axes are predicted directly in the gravity-aligned, view-relative frame. Revolute limits are expressed in radians, while prismatic displacements are normalised by $s$. Prismatic joints require no pivot.

\paragraph{Scene inputs.}
\label{sec:supp_view_scene}
The \texttt{objects} field groups region identifiers by object, with the static body listed first. A joint attached to \texttt{base} refers to its own object's static body. The designated reference object defines the observation centre and scale, and all other objects share that frame. Object membership and body grouping therefore yield one hierarchy per object within a jointly generated scene structure, using direct autoregressive prediction without per-instance optimisation.

\subsection{Supervision Construction}
\label{sec:supp_view_supervision}
Source meshes and joints are first converted to consistent coordinates and units. We place each asset at rest using forward kinematics, then render aligned RGB, part-ID, and depth images from one canonical and several sampled cameras. Each camera gives a separate static training example. Joint axes and pivots are transformed into the observation frame, and pivots are projected to obtain image-space targets. For motion examples, a clip actuates one joint while retaining the reference inventory and frame. Scene examples additionally record the membership of all annotated objects.

Fixed-joint chains are collapsed into rigid bodies before serialisation, retaining one articulated attachment per moving body. This prevents a handle fixed to a door, or two fragments of the same door, from receiving independent copies of the hinge. Meshless scaffold links are removed without discarding fixed attachments between physical parts.

To vary part granularity, split augmentation divides a selected articulated leaf part by a plane through its centroid, while merge augmentation combines sibling parts with compatible articulation. The segmentation is re-rendered and the grouped target rebuilt, so fragments moving together share one joint. These operations vary the decomposition while preserving its intended kinematics.

\subsection{Training Mixture and Configuration}
\label{sec:supp_view_training}
\label{sec:supp_view_corpus}

\paragraph{Training mixture.}
Our training pool comprises PhysX-Mobility~\citep{cao2025physx} ($1{,}633$ objects, $13{,}064$ views), its decomposition variants ($4{,}561$ variants, $36{,}488$ views), PhysXNet~\citep{cao2025physx} ($40{,}000$ objects, $160{,}000$ views), GRScenes~\citep{grutopia} objects ($2{,}502$ objects, $10{,}008$ views), and $2{,}338$ GRScenes room views. To balance source contributions, each epoch samples $50\%$ of the decomposition-variant pool and $40\%$ of the PhysXNet pool, while retaining all examples from the remaining sources. These sampling rates apply to the individual source pools, rather than their proportions in the final mixture. We additionally include $5{,}954$ PhysX-Mobility articulation clips, yielding approximately $114$k examples per epoch.

\paragraph{Training configuration.}
We adapt Qwen3-VL-4B-Instruct using decoder LoRA with rank $64$ and scaling factor $\alpha=128$, jointly training the vision--language merger while freezing the vision encoder and base decoder weights. The merger weights are saved alongside the adapters. We minimise cross-entropy over answer tokens for two epochs using bfloat16 precision and an effective batch size of $32$, distributed across four workers with one example per worker and eight gradient accumulation steps. Training uses a cosine learning-rate schedule with a peak of $10^{-4}$, a warmup ratio of $0.03$, zero weight decay, and a gradient-clipping threshold of $1$. The context length is limited to $8192$ tokens, and the image budget is $65{,}536$ pixels. We enable gradient checkpointing and ZeRO stage~2.

\section{Evaluation Protocols}
\label{sec:supp_eval}
\label{sec:supp_exp_protocols}

\subsection{Datasets and Observation Settings}
\label{sec:supp_splits}

\paragraph{Objects.}
PhysX-Mobility contains $387$ test objects. SimReady contains $242$ kitchen assets from $14$ categories and is absent from training; its categories overlap PartNet-Mobility, so it tests transfer across instances and asset sources. GRScenes contains $3{,}099$ converted movable assets: $2{,}502$ training objects, $306$ test instances from seen categories, and $291$ objects in seven categories excluded from its training partition: hearth, electric cooker, TV stand, tea table, sideboard, chest of drawers, and shoe cabinet. Object comparisons use one reference view unless otherwise stated. PhysX-Anything and PhysX-Omni transfer metrics are conditional on objects producing an output.

\paragraph{Scenes.}
We render $2{,}897$ views in $69$ GRScenes-100 rooms, each camera placed in front of a target object while also annotating neighbouring articulated objects. Views contain approximately $2.2$--$2.9$ objects and $5$--$7$ joints on average across splits. There are $2{,}338$ training views, $305$ seen-category test views, and $254$ held-out-category test views. Training views contain only training models; test views are partitioned by the target object, without imposing the same partition on neighbours. In held-out-target views, $74\%$ of neighbours are training models and $81\%$ of views mix partitions. In seen-target views, $12\%$ of neighbours belong to held-out categories. These splits therefore test prediction around different target categories, not wholly unseen rooms or exclusively unseen objects.

The main scene table pools all $559$ test views and retains out-of-frame pivots. For PhysX-Omni, we run each articulated object separately with the background masked and place its predictions back into the scene; ViewStruct processes the room view jointly. 

\paragraph{Real videos.}
Arti4D contains $342$ interactions in four indoor scenes, comprising $228$ prismatic and $114$ revolute joints. ViewStruct receives an initial RGB-D frame with Mobile-SAM regions and Kinect depth, optionally supplemented by two RGB frames and a $16$-frame interaction clip as specified in \Cref{tab:arti4d}. It uses no real-data adaptation or temporal camera poses. ArtiPoint receives the full posed RGB-D sequence and reference interaction cues. For visualisation only, camera poses reproject predicted axes into successive frames, and depth determines which segments are drawn as occluded.

\subsection{Baselines and Part Correspondence}
\label{sec:supp_rebinding}

\paragraph{Inputs.}
Zero-shot Qwen3-VL receives the same images, region descriptions, and output schema as ViewStruct. PhysX-Anything~\citep{cao2025physx} and PhysX-Omni~\citep{cao2026physx} infer joints from the image through their VLM stages; geometry generation is skipped, and reference meshes are used for association with the evaluation inventory, not as visual input to those stages. URDF-Anything+~\citep{li2025urdf} receives the image and a whole-object mesh assembled from reference parts. We use its released checkpoint without shell thickening, which removed articulated components in our tests. Particulate~\citep{li2026particulate} receives the merged reference mesh without an image and predicts its own decomposition and joints. OPDMulti~\citep{sun2024opdmulti} receives the reference RGB image alone and predicts a mask, motion type, axis, and origin per openable part; the released motion-state checkpoint additionally predicts the joint state, from which a motion range follows. Its masks are matched to our inventory by intersection over union against the reference segmentation, so no geometric re-binding is required, and its camera-frame axes and origins are mapped into the observation frame with the view pose. It receives no camera intrinsics, so its absolute depth follows its training camera: on the room renders the predicted pivot depth is $0.55\times$ the reference value, and we remove this single factor per view by placing the predicted origins at the object's known distance, which raises scene success from $10.1$ to $11.8$. On Arti4D, recorded with a camera close to its own, the same correction is harmful ($0.42\rightarrow0.53$\,m) and predictions are reported unmodified.
SINGAPO~\citep{liu2025singapo} is evaluated on the $65$ StorageFurniture objects supported by its category vocabulary. Of these, $52$ overlap its training set. Particulate has $81$ training-overlap objects among the $387$ PhysX-Mobility test objects. These overlaps are marked in the detailed comparison and preclude treating those rows as fully held-out evaluations.

\paragraph{URDF conversion and rebinding.}
We compose joint transforms through the predicted URDF, remove meshless scaffold links, and express axes, pivots, and part geometry in the reference scoring frame. ViewStruct outputs already use the reference part identifiers. For generated decompositions requiring association, a Hungarian assignment matches normalised centroids $c_p$ and extents $e_p$ using
\begin{equation}
    C(p,q)=\lVert c_p-c_q\rVert_2+0.5\lVert e_p-e_q\rVert_2.
\end{equation}
Unmatched predicted nodes retain distinct identifiers and count as false positives when they carry joints. This association is applied to every method that produces its own decomposition, and it matters: the reference identifiers are assigned in a shuffled order, so pairing predictions by identifier instead of geometry attaches each joint to a different part and penalises correct parameters. On PhysX-Mobility, geometric association raises articulation success from $74.7$ to $81.6$ for PhysX-Anything, $77.8$ to $85.3$ for PhysX-Omni, $46.3$ to $68.2$ for Particulate, and $8.6$ to $14.0$ for URDF-Anything+, and it reduces the mean pivot error of the first two by more than a factor of two. Repeated identical parts are where identifier pairing fails most clearly: on an oven with seven knobs, every predicted axis, pivot and range of PhysX-Omni is correct, yet each joint is attached to a neighbouring knob, giving zero successes before association and eight of eight after. These policies are part of the evaluation and do not assume identical decompositions across methods.

\subsection{Metrics and Normalisation}
\label{sec:supp_metrics}

Predictions are paired with reference examples by identifier and expanded into parent maps and joints. Parse success is reported separately; structural and joint metrics are aggregated over parsed predictions. Thus, conditional accuracy must be read together with parse success and joint coverage.

\paragraph{Connectivity.}
Parent accuracy measures agreement over reference parts after star-anchor canonicalisation of reference instance groups, applying the same smallest-ID anchor convention to both graphs. Edge $F_1$ compares undirected parent edges, removing sensitivity to edge orientation. The rerooted diagnostic additionally roots each connected component at its smallest identifier. Connectivity scores for generated decompositions combine hierarchy accuracy with coverage of the reference inventory. On PhysX-Mobility, PhysX-Anything, PhysX-Omni, Particulate, and URDF-Anything+ cover $85.5\%$, $88.0\%$, $48.1\%$, and $24.8\%$ of reference parts, respectively, compared with $98.9\%$ for ViewStruct. 
% Restricting parent accuracy to matched parts raises PhysX-Anything and PhysX-Omni from $45.4\%$ and $45.8\%$ to $52.4\%$ and $51.1\%$, illustrating the effect of coverage.

% Flat reference hierarchies make connectivity weakly informative. An all-static prediction with no joints achieves parent/edge scores of $90.5/88.5\%$ on SimReady, $99.7/99.5\%$ on held-out GRScenes objects, and $99.9/99.9\%$ on scene views. The same prediction scores $56.0/24.6\%$ on PhysX-Mobility, which contains nested attachments. We therefore retain connectivity in the main PhysX-Mobility table and omit it from the main SimReady and GRScenes tables. Additional diagnostic tables retain these columns for completeness.

\paragraph{Joint presence and geometry.}
Joint-presence precision, recall, and $F_1$ compare which corresponding parts carry an articulated joint, independently of whether its parameters are correct. Type accuracy measures revolute/prismatic agreement on matched joints. Axis accuracy uses the undirected angle $\arccos|\hat a_{\mathrm{pred}}^{\top}\hat a_{\mathrm{gt}}|$ with threshold $0.25$\,rad ($14.3^\circ$). Range IoU compares predicted and reference motion intervals.

\begin{figure}[htbp]
    \centering
    \includegraphics[width=\linewidth]{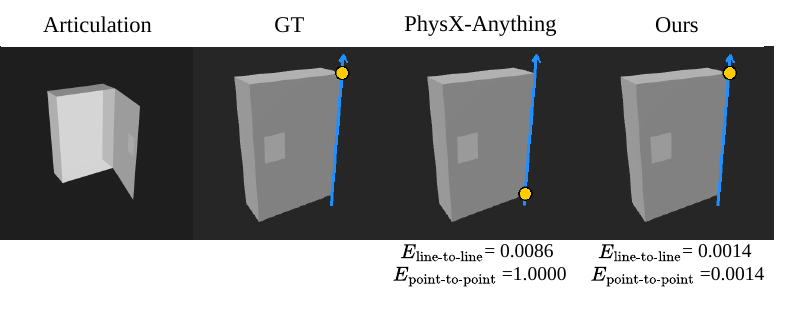}
    \caption{\textbf{Hinge-pivot equivalence.} Displacement along a shared axis leaves the rotation unchanged; the scoring convention is given in \Cref{eq:supp_pivot_error}.}
    \label{fig:hinge}
\end{figure}

\paragraph{Hinge-position error.}
\label{sec:supp_hinge_origin}
For a pivot $x$ and its axis line $\ell$, the object and scene evaluations use
\begin{equation}
    d_{\mathrm{pivot}}=
    \max\!\left(d(x_{\mathrm{gt}},\ell_{\mathrm{pred}}),
                d(x_{\mathrm{pred}},\ell_{\mathrm{gt}})\right).
    \label{eq:supp_pivot_error}
\end{equation}
This symmetric point-to-line error differs from the shortest distance between two infinite lines. In particular, intersecting but misoriented axes can have zero shortest-line distance while failing this criterion. When the directions agree, shifting a pivot along its axis does not affect the error (\Cref{fig:hinge}); this invariance does not hold in general when the axes differ. Errors are averaged over matched hinges and normalised by the object's largest extent. Prismatic origins are not scored because translation is independent of pivot position. Scene joints are transformed from the shared prediction frame into their own object's normalised frame before scoring.

\paragraph{Articulation success.}
Following the failure ordering of \citet{le2024articulate}, a matched joint is assigned to its first failed gate: type, axis, origin, then limit. The thresholds are $0.25$\,rad for the axis, $0.05$ object-extent units for the hinge pivot, and $1.6057$ for the difference between range spans (radians for hinges; the scorer uses the same numerical threshold in normalised displacement units for slides). Slides skip the origin gate. Matched-joint success is the fraction passing all gates. The scene table also divides successful matches by all reference joints in the scored rows, counting unmatched joints as failures; parse failures remain separately reported. Manipulation uses its own all-reference-joint denominator, including missing predictions as failures.

The limit gate is loose, so range IoU provides a more informative measure of limit estimation. Range annotations also vary in quality: PhysX-Mobility and SimReady have $152$ and $77$ distinct hinge limits, with modal spans covering $38\%$ and $30\%$ of hinges. In contrast, $96\%$ of GRScenes hinge limits use the converter's $90^\circ$ default. GRScenes hinge-range agreement therefore largely measures agreement with this default. Prismatic ranges vary across sources.

\paragraph{Real-video metrics.}
For Arti4D, axis error is the undirected angle in degrees, and hinge-position error follows the benchmark's common-perpendicular axis distance in metres. This position metric differs from \Cref{eq:supp_pivot_error}. We report means and medians, and the number of evaluated outputs for each configuration.

\subsection{Inference-Time Measurement}
\label{sec:supp_timing}
All methods are timed sequentially on a dedicated A100-40GB using the same three objects, selected at the $25$th, $50$th, and $75$th percentiles of part count among $343$ commonly runnable objects (two, three, and six parts). One-off loading is excluded for ViewStruct, PhysX-Anything, and PhysX-Omni, which retain their models in memory. Loading is included for URDF-Anything+ and Particulate, which launch a process per object. PhysX-Anything and PhysX-Omni timings cover only their VLM joint-prediction stages and exclude geometry generation; they are not full reconstruction costs.

\section{Additional Results}
\label{sec:supp_results}

\subsection{Detailed Object Comparisons}
\Cref{tab:physx_detailed} expands the main PhysX-Mobility comparison with parse rates, failure attribution, and the category-restricted SINGAPO comparison. The ViewStruct configuration without GRScenes training is included as a supplementary diagnostic of the training mixture.

\begin{table}[t]
\centering
\caption{\textbf{Detailed PhysX-Mobility comparison.} Full-set and StorageFurniture-subset diagnostics. $^\dagger$SINGAPO: $52/65$ training-overlap objects; $^\P$Particulate: $81/387$. Scores are percentages except normalised pivot error. Part-matching conventions are described in the text.}
\label{tab:physx_detailed}
\footnotesize
\setlength{\tabcolsep}{3.2pt}
\resizebox{\linewidth}{!}{%
\begin{tabular}{@{}c@{}}
\begin{tabular}{@{}l@{\hspace{4pt}}ccc@{\hspace{6pt}}ccccc@{}}
\toprule
& \multicolumn{3}{c}{\textit{Structure}} & \multicolumn{5}{c}{\textit{Joint}} \\
\cmidrule(lr){2-4} \cmidrule(lr){5-9}
& Parse$\uparrow$ & Parent$\uparrow$ & Edge $F_1\uparrow$ & Pres. $F_1\uparrow$ & Type$\uparrow$ & Axis$\uparrow$ & Pos.$\downarrow$ & Range IoU$\uparrow$ \\
\midrule
Qwen3-VL-4B (zero-shot) & 81.4 & 19.2 & 19.9 & 31.5 & 66.0 & 41.2 & 0.391 & 30.5 \\
PhysX-Anything & \textbf{100.0} & 61.0 & 67.9 & 85.9 & 96.2 & 94.4 & 0.118 & 26.3 \\
PhysX-Omni & \textbf{100.0} & 70.1 & 73.6 & 85.9 & \textbf{97.8} & \textbf{95.8} & 0.096 & 26.5 \\
URDF-Anything+ & 99.2 & 17.3 & 14.7 & 18.6 & 73.0 & 58.5 & 0.269 & 34.1 \\
Particulate$^\P$ & \textbf{100.0} & 37.8 & 45.3 & 61.0 & 97.5 & 94.8 & \textbf{0.036} & 54.1 \\
ViewStruct (PhysX only) & 98.7 & 91.1 & 91.4 & \textbf{97.2} & 96.2 & 86.5 & 0.087 & 69.8 \\
ViewStruct & 99.7 & \textbf{92.3} & \textbf{92.7} & 96.8 & 96.6 & 89.9 & 0.084 & \textbf{72.3} \\
ViewStruct (MoGe depth) & 99.5 & \textbf{92.3} & 91.0 & 96.5 & 95.6 & 89.3 & 0.094 & 69.1 \\
\midrule
\multicolumn{9}{@{}l}{\textit{StorageFurniture subset} ($65$ objects; the only category shared with SINGAPO's taxonomy)} \\
SINGAPO$^\dagger$ & \textbf{100.0} & 29.8 & 30.4 & 51.2 & 84.1 & 86.6 & 0.569 & 45.5 \\
PhysX-Anything & \textbf{100.0} & 38.4 & 26.8 & 79.8 & 90.2 & 79.5 & 0.190 & 18.5 \\
Particulate$^\P$ & \textbf{100.0} & 32.7 & 27.8 & 96.4 & \textbf{100.0} & \textbf{100.0} & \textbf{0.014} & \textbf{93.5} \\
ViewStruct & \textbf{100.0} & \textbf{87.8} & \textbf{86.4} & \textbf{98.2} & 98.2 & 96.3 & 0.036 & 82.7 \\
\bottomrule
\end{tabular} \\
\noalign{\vskip 18pt}
\begin{tabular}{@{}l@{\hspace{6pt}}cccccc@{}}
\toprule
& \multicolumn{6}{c}{\textit{Articulation failure attribution}} \\
\cmidrule(lr){2-7}
& Type$\downarrow$ & Axis$\downarrow$ & Origin$\downarrow$ & Limit$\downarrow$ & Success$\uparrow$ & Succ.\,obj.$\uparrow$ \\
\midrule
Qwen3-VL-4B (zero-shot) & 34.0 & 35.6 & 28.7 & 0.6 & 1.1 & 1.0 \\
PhysX-Anything & 3.8 & \textbf{3.0} & 10.7 & 0.9 & 81.6 & 40.2 \\
PhysX-Omni & \textbf{2.2} & 3.1 & 9.1 & \textbf{0.3} & \textbf{85.3} & 38.9 \\
URDF-Anything+ & 27.0 & 25.0 & 33.0 & 1.0 & 14.0 & 2.3 \\
Particulate$^\P$ & 2.5 & 3.9 & \textbf{4.5} & 20.9 & 68.2 & 51.0 \\
ViewStruct (PhysX only) & 3.8 & 11.1 & 8.1 & 0.6 & 76.5 & \textbf{53.0} \\
ViewStruct & 3.4 & 8.1 & 6.2 & 0.8 & 81.5 & 52.5 \\
ViewStruct (MoGe depth) & 4.4 & 8.0 & 8.6 & 0.7 & 78.3 & 48.7 \\
\bottomrule
\end{tabular}
\end{tabular}}
\end{table}

\Cref{tab:grscenes_seen} evaluates $306$ unseen instances from GRScenes categories represented during training. ViewStruct achieves $70.6\%$ matched-joint success, compared with $48.7\%$ on the held-out-category objects in the main paper. The two splits distinguish transfer to new instances from transfer to categories excluded from the GRScenes training partition. Their nearly flat hierarchies limit the informativeness of the connectivity columns.

\begin{table}[t]
\centering
\small
\setlength{\tabcolsep}{5.7pt}
\caption{\textbf{GRScenes objects from seen categories} ($306$ test instances). Single-view results; percentages except normalised pivot error.}
\label{tab:grscenes_seen}
\begin{tabular}{@{}lrrrrrrrrr@{}}
\toprule
Method & Parent$\uparrow$ & Edge $F_1\uparrow$ & Joint $F_1\uparrow$ & Type$\uparrow$ & Axis$\uparrow$ & Pivot$\downarrow$ & Range$\uparrow$ & Success$\uparrow$ & Succ.\,obj.$\uparrow$ \\
\midrule
Qwen3-VL-4B (zero-shot) & 56.0 & 52.3 & 55.3 & 69.0 & 42.9 & 0.352 & 37.2 & 0.0 & 0.0 \\
URDF-Anything+ & 48.6 & 47.5 & 46.5 & 74.5 & 47.0 & 0.443 & 37.0 & 0.5 & 0.4 \\
Particulate & 82.9 & 81.2 & 87.0 & 88.1 & 80.1 & 0.150 & 79.9 & 63.7 & 57.2 \\
ViewStruct & \textbf{99.7} & \textbf{99.9} & \textbf{99.8} & \textbf{93.2} & \textbf{88.6} & \textbf{0.100} & \textbf{86.1} & \textbf{70.6} & \textbf{66.3} \\
\bottomrule
\end{tabular}
\end{table}

\subsection{Ablations}
\label{sec:supp_ablation}
Two components of the recipe are isolated by retraining the reported model with a single change and evaluating on
the same PhysX-Mobility single-view split. Everything else -- backbone, adapters, optimisation, training mixture,
target format and the number of steps -- is held fixed, so the differences in \Cref{tab:supp_ablation} are
attributable to the change alone.

\paragraph{Observation frame.}
Replacing the gravity-aligned frame of \Cref{eq:frame_basis} with the raw camera frame, leaving the origin and unit
length unchanged, costs $3.9$ points of articulation success. The loss is concentrated in the axis channel
($89.9\rightarrow86.6$) and in the success rate that depends on it, while joint presence is unchanged
($96.8\rightarrow96.9$) and the hinge pivot is unaffected ($0.084\rightarrow0.082$). This is the expected pattern:
pivots are read off the image and do not depend on the 3D convention, whereas an axis expressed in camera
coordinates changes with the camera's pitch and roll, so the same vertical hinge takes a different target value in
every view and the model must absorb that variance. Aligning one axis of the frame with gravity removes it for the
vertical hinges that dominate the corpus; horizontal directions remain view-dependent in both frames.

\paragraph{Decomposition augmentation.}
Removing the split-merge rows, which vary the supplied part granularity while preserving the intended kinematics,
measures how much of the model's robustness comes from having seen alternative decompositions of the same object
rather than from the object inventory itself. It is the larger of the two effects: articulation success drops by
$10.0$ points ($81.5\rightarrow71.5$) although the same objects and views remain in the training set. The joints
are still found ($96.8\rightarrow96.2$ presence), but their parameters degrade across every channel, the axis most
($89.9\rightarrow82.8$), followed by the motion range ($72.3\rightarrow66.3$) and the hinge pivot
($0.084\rightarrow0.096$). Grouping is also weaker ($92.3\rightarrow88.6$ parent), which is what the augmentation
targets directly: without rows in which a part is split or two parts are merged, the model associates a particular
segmentation of an object with a particular answer instead of inferring which supplied regions move together.

\paragraph{Error in the gravity direction.}
The model receives no numerical gravity: $c$, $s$ and the depth intervals of \Cref{eq:prompt-camnative} are read
from the silhouette and the depth map, and the prompt describes the frame only in words. Gravity therefore enters
at a single point, when the output is interpreted, so an inertial sensor that is wrong by $\delta$ is exactly
equivalent to rotating every predicted axis and pivot by $\delta$ before comparing them with the reference. We
apply that rotation about a random horizontal axis, per object, and rescore; no retraining is involved.
\Cref{tab:gravity} shows the result. Errors up to $2^\circ$ are free, $5^\circ$ costs $1.7$ points, and the
degradation becomes severe only as $\delta$ approaches the $14.3^\circ$ axis threshold, which it then consumes.
An inertial sensor is an order of magnitude more accurate than that, and monocular gravity estimators sit near
$2$--$3^\circ$; joint type, which does not depend on the frame, is unaffected throughout.

\begin{table}[!htb]
\centering
\small
\setlength{\tabcolsep}{6pt}
\caption{\textbf{Cost of a wrong gravity direction} (PhysX-Mobility, $387$ objects, single view). The reported
predictions rescored as if the assumed gravity were tilted by $\delta$ about a random horizontal axis.}
\label{tab:gravity}
\begin{tabular}{@{}lrrrrr@{}}
\toprule
Gravity error & Type$\uparrow$ & Axis$\uparrow$ & Pivot$\downarrow$ & Success$\uparrow$ & Succ.\,obj.$\uparrow$ \\
\midrule
none & 96.6 & 89.9 & 0.084 & 81.5 & 52.5 \\
$1^\circ$ & 96.6 & 90.0 & 0.084 & 81.6 & 52.5 \\
$2^\circ$ & 96.6 & 90.0 & 0.084 & 81.5 & 52.2 \\
$5^\circ$ & 96.6 & 88.5 & 0.087 & 79.8 & 52.5 \\
$10^\circ$ & 96.6 & 82.5 & 0.096 & 70.7 & 45.7 \\
$20^\circ$ & 96.6 & 29.8 & 0.120 & 25.6 & 8.3 \\
\bottomrule
\end{tabular}
\end{table}

\begin{table}[!htb]
\centering
\footnotesize
\setlength{\tabcolsep}{5pt}
\caption{\textbf{Ablations on PhysX-Mobility} ($387$ objects, single view, strict origin metric). Each row retrains
the reported recipe with one change. Metrics and units follow \Cref{tab:physx_anything}.}
\label{tab:supp_ablation}
\begin{tabular}{@{}lrrrrrrrrr@{}}
\toprule
Recipe & Parent$\uparrow$ & Edge $F_1\uparrow$ & Joint $F_1\uparrow$ & Type$\uparrow$ & Axis$\uparrow$ & Pivot$\downarrow$ & Range$\uparrow$ & Success$\uparrow$ & Succ.\,obj.$\uparrow$ \\
\midrule
ViewStruct & \textbf{92.3} & \textbf{92.7} & 96.8 & \textbf{96.6} & \textbf{89.9} & 0.084 & \textbf{72.3} & \textbf{81.5} & 52.5 \\
\quad camera frame instead of gravity-aligned & 91.1 & 90.6 & \textbf{96.9} & 96.1 & 86.6 & \textbf{0.082} & 65.7 & 77.6 & \textbf{54.5} \\
\quad without decomposition augmentation & 88.6 & 88.4 & 96.2 & 94.8 & 82.8 & 0.096 & 66.3 & 71.5 & 50.4 \\
\bottomrule
\end{tabular}
\end{table}

\subsection{Zero-Shot Backbone Scaling}
\label{sec:zero_shot_scaling}
We compare zero-shot Qwen3-VL-4B and Qwen3-VL-8B on the same $387$ PhysX-Mobility objects, using identical inputs, output schema, greedy decoding, and maximum output length. Scaling the backbone improves parse success from $81.4\%$ to $100\%$ and joint-presence $F_1$ from $31.5\%$ to $69.4\%$, but matched-joint success remains low ($1.1\%$ and $0.6\%$). ViewStruct achieves $81.5\%$ under the same evaluation. Thus, increasing backbone size alone does not close the articulation gap in this setting. Per-joint accuracies should be interpreted with coverage: the two zero-shot models recover different subsets of reference joints.

\begin{table}[t]
\centering
\small
\setlength{\tabcolsep}{6pt}
\begin{tabular}{lccc}
\toprule
                                      & Qwen3-VL-4B & Qwen3-VL-8B & ViewStruct \\
                                      & (zero-shot) & (zero-shot) & (single view) \\
\midrule
parse\,success$\uparrow$ & 81.4\% & 100.0\% & \textbf{99.7\%} \\
\midrule
\multicolumn{4}{l}{\textit{Structure}} \\
parent\,acc\,(canon)$\uparrow$ & 19.2\% & 38.7\% & \textbf{92.3\%} \\
parent\,acc\,(rerooted)$\uparrow$ & 22.3\% & 29.3\% & \textbf{92.6\%} \\
edge\,$F_1\uparrow$ & 19.9\% & 27.3\% & \textbf{92.7\%} \\
\midrule
\multicolumn{4}{l}{\textit{Joint}} \\
joint\,presence\,$F_1\uparrow$ & 31.5\% & 69.4\% & \textbf{96.8\%} \\
joint\,type\,acc$\uparrow$ & 66.0\% & 41.2\% & \textbf{96.6\%} \\
joint\,axis\,acc$\uparrow$ & 41.2\% & 32.9\% & \textbf{89.9\%} \\
joint\,position\,err$\downarrow$ & 0.391 & 0.324 & \textbf{0.084} \\
joint\,range\,IoU$\uparrow$ & 30.5\% & 21.5\% & \textbf{72.3\%} \\
\midrule
articulation success$\uparrow$ & 1.1\% & 0.6\% & \textbf{81.5\%} \\
articulation success per object$\uparrow$ & 1.0\% & 1.3\% & \textbf{52.5\%} \\
\bottomrule
\end{tabular}
\caption{\textbf{Backbone scaling versus adaptation.} Zero-shot Qwen3-VL and ViewStruct on $387$ PhysX-Mobility objects with identical input and output formats.}
\label{tab:zero_shot_scaling}
\end{table}

\subsection{Visibility and Scene Diagnostics}
\label{sec:supp_visible}
\Cref{tab:visible_only} excludes joints on parts with no pixels in the reference segmentation. On the canonical PhysX-Mobility view, $118$ of $1{,}879$ reference joints ($6.3\%$) belong to hidden parts. Restricting to visible parts raises ViewStruct success from $81.5\%$ to $83.4\%$ and yields similarly modest changes for the baselines. The separate video row uses $2{,}172$ clips and scores the moving part, which is treated as visible throughout the diagnostic.

The URDF-Anything+ row is retained from the separate visibility scoring export, whose parse rate is $82.4\%$, versus $99.2\%$ in the main comparison export. Its all-part score is therefore not the same result as the main-table row; the all/visible pair should be read within this diagnostic.

\begin{table}[htbp]
\centering
\caption{\textbf{Visibility diagnostic on PhysX-Mobility.} All-part and visible-part scores (\%). The video row uses a separate clip set; the URDF-Anything+ export differs from the main comparison, as detailed in the text.}
\label{tab:visible_only}
\small
\setlength{\tabcolsep}{6pt}
\begin{tabular}{@{}lcccccccc@{}}
\toprule
& \multicolumn{2}{c}{Success$\uparrow$} & \multicolumn{2}{c}{Success obj.$\uparrow$} & \multicolumn{2}{c}{Axis$\uparrow$} & \multicolumn{2}{c}{Origin fail$\downarrow$} \\
\cmidrule(lr){2-3} \cmidrule(lr){4-5} \cmidrule(lr){6-7} \cmidrule(lr){8-9}
& all & visible & all & visible & all & visible & all & visible \\
\midrule
Qwen3-VL-4B (zero-shot) & 1.1 & 1.2 & 1.0 & 1.0 & 41.2 & 41.7 & 28.7 & 30.1 \\
PhysX-Anything & 81.6 & 82.1 & 40.2 & 36.9 & 94.4 & 94.5 & 10.7 & 10.5 \\
PhysX-Omni & \textbf{85.3} & \textbf{86.0} & 38.9 & 35.3 & \textbf{95.8} & \textbf{96.0} & 9.1 & 9.0 \\
URDF-Anything+ & 14.0 & 11.4 & 2.3 & 1.8 & 58.5 & 57.7 & 33.0 & 34.9 \\
Particulate (GT mesh) & 68.2 & 68.9 & 51.0 & 46.3 & 94.8 & 95.4 & \textbf{4.5} & \textbf{5.0} \\
ViewStruct & 81.5 & 83.4 & \textbf{52.5} & \textbf{54.6} & 89.9 & 91.9 & 6.2 & 6.2 \\
ViewStruct (MoGe depth) & 78.3 & 79.7 & 48.7 & 50.3 & 89.3 & 91.1 & 8.6 & 8.9 \\
\midrule
ViewStruct, video (moving part) & 54.4 & 54.4 & 41.2 & 41.2 & 73.6 & 73.6 & 16.6 & 16.6 \\
\bottomrule
\end{tabular}
\end{table}

\subsection{Qualitative Results}
\label{sec:supp_qual}
Across the object and scene visualisations, hinged parts rotate by up to $45^\circ$ about the predicted line and sliding parts translate along their predicted axis and range. The moved geometry is composited with the rest-pose geometry under occlusion. Red denotes hinged motion, blue sliding motion, and yellow the axes; dimmed panels indicate missing outputs. These renders visualise inferred kinematics, rather than physically simulated contact. The Claude Fable 5.1 panels use a Claude Code agent with image-file access and are distinct from the quantitative zero-shot Qwen3-VL baseline.

\begin{figure}[htbp]
    \centering
    \includegraphics[width=\linewidth]{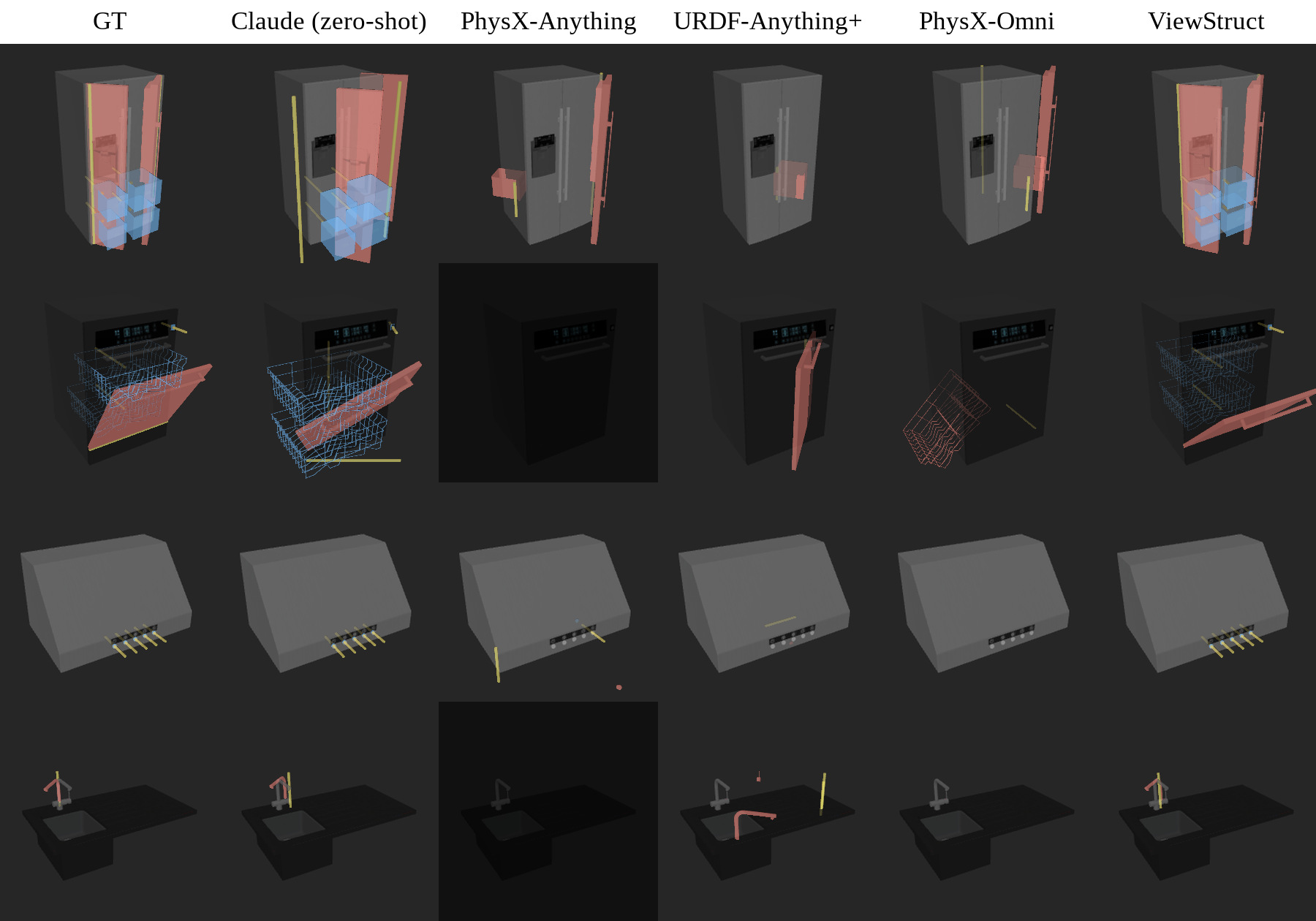}
    \caption{\textbf{Articulation predictions on SimReady.} The image baselines of \Cref{fig:comparison} on an unseen asset source. Red: hinged motion; blue: sliding motion; yellow: axes. Dimmed panels indicate missing outputs.}
    \label{fig:qual_generalisation}
\end{figure}

\begin{figure}[htbp]
    \centering
    \includegraphics[width=\linewidth]{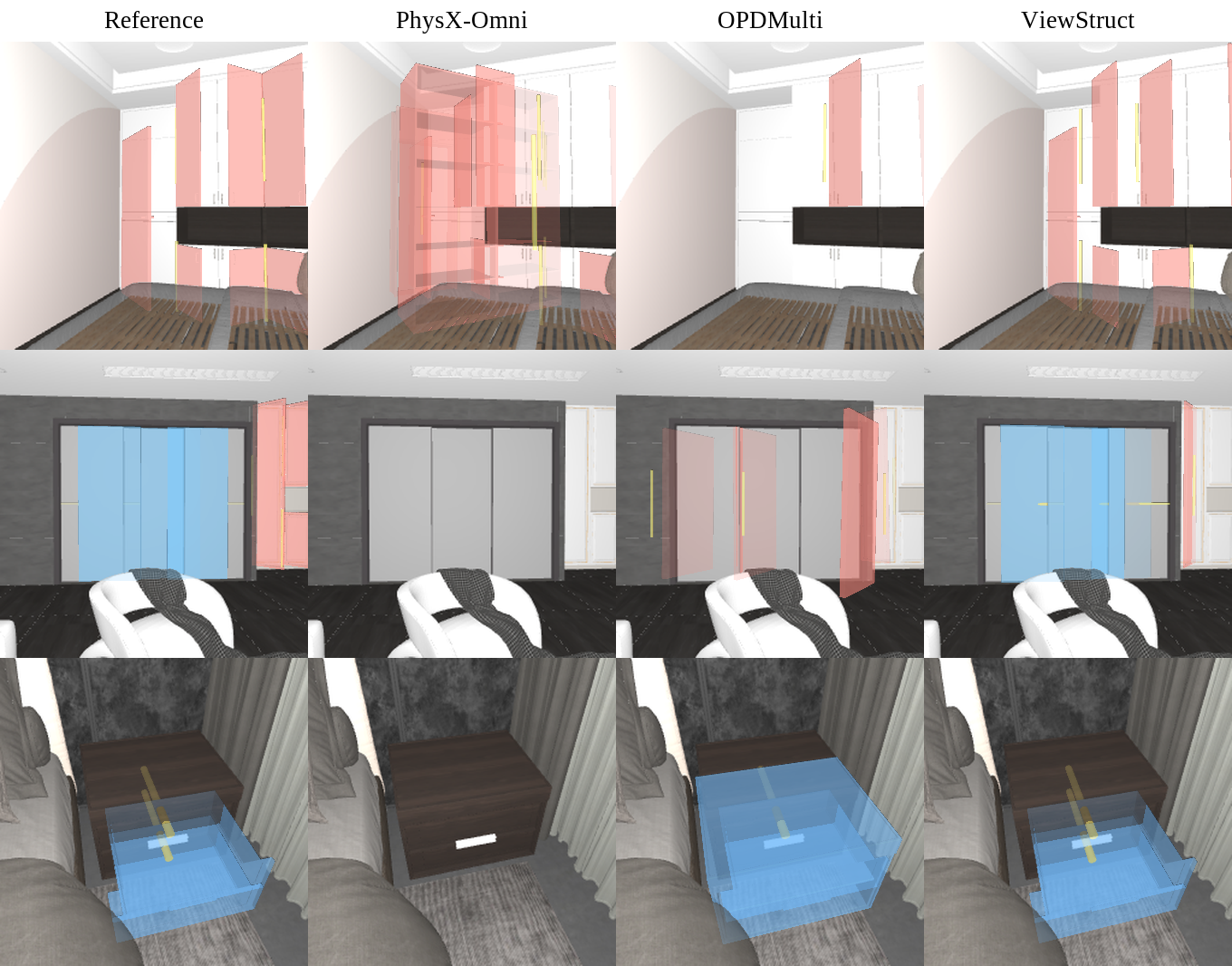}
    \caption{\textbf{Additional GRScenes scene-view predictions.} Reference, PhysX-Omni, OPDMulti and ViewStruct on the same views, with the same panels as \Cref{fig:scene_qual}. Each method's own joints move the same reference meshes.}
    \label{fig:qual_scenes}
\end{figure}

\Cref{fig:qual_generalisation} shows transfer to SimReady. ViewStruct recovers refrigerator doors and drawers, range-hood buttons, and a faucet swivel, while the examples also expose missed joints and incorrect hinge edges in baseline predictions. PhysX-Anything's visual examples come from a $56$-object SimReady subset. \Cref{fig:qual_scenes} shows multi-object grouping and articulation in GRScenes: a room door with six wardrobe hinges, sliding wardrobe doors beside a hinged door, and a bedside cabinet whose drawer slides out. In the last view PhysX-Omni leaves the cabinet static while OPDMulti translates the whole body instead of the drawer, illustrating how errors in grouping propagate into the predicted motion.

\section{From Predicted Joints to Manipulation Trajectories}
\label{sec:supp_downstream}

ViewStruct provides motion constraints for generating robot end-effector trajectories. Given a contact point on a predicted moving body, the joint type determines the trajectory shape: a prismatic joint defines a straight path, while a revolute joint defines a circular arc. Persistent part identifiers associate each joint with the geometry to be manipulated.

\paragraph{Mapping predictions to the robot frame.}
We first transform the predicted joint parameters from the normalised view-relative frame into the robot planning frame. Let $R$ map directions from the prediction frame to the planning frame, $\mathbf{c}$ denote the prediction-frame origin in planning coordinates, and $s$ denote the normalisation scale. The joint pivot and axis become
\begin{equation}
    \mathbf{o}=\mathbf{c}+sR\mathbf{o}_v,
    \qquad
    \hat{\mathbf{a}}=R\hat{\mathbf{a}}_v.
\end{equation}
Prismatic limits are multiplied by $s$, while angular limits remain unchanged.

\paragraph{Generating the motion path.}
Let $\mathbf{q}_0$ be the initial contact position in the planning frame. For a prismatic joint, the target position follows
\begin{equation}
    \mathbf{q}(d)=\mathbf{q}_0+d\hat{\mathbf{a}},
\end{equation}
where $d$ is the displacement from the initial configuration. For a revolute joint, the target follows
\begin{equation}
    \mathbf{q}(\theta)
    =\mathbf{o}
    +\operatorname{Rot}(\hat{\mathbf{a}},\theta)
    (\mathbf{q}_0-\mathbf{o}),
\end{equation}
where $\operatorname{Rot}(\hat{\mathbf{a}},\theta)$ denotes rotation by $\theta$ about the predicted axis. Sampling $d$ or $\theta$ over the intended motion produces a sequence of Cartesian waypoints, with travel constrained by the predicted joint limits.

These waypoints provide the positional reference for robot motion planning. Combined with a chosen grasp orientation, they can be converted into robot joint configurations through inverse kinematics and executed by a controller. ViewStruct therefore supplies the articulation-dependent path, while grasp selection, collision avoidance, and trajectory timing are handled downstream.

\subsection{Compliant Manipulation on PhysX-Mobility}
\label{sec:supp_manip_protocol}

\paragraph{Simulation setup.}
We reconstruct each object in PyBullet using its reference joints and limits, with a fixed base, gravity disabled, and no joint motors. A compliant gripper is represented by a force-bounded point constraint attached at a supplied contact location: the mesh point farthest from the reference axis for a hinge, or the moving part's centre for a slide. These privileged contacts and simplified physics isolate articulation prediction quality; grasp discovery and collision avoidance are not evaluated.

\paragraph{Trajectories and success criteria.}
The gripper follows the predicted joint associated with the contacted part. For a hinge, it traces a $45^\circ$ arc around the predicted axis and pivot. For a slide, it moves along the predicted direction by the smaller of the predicted range and $0.3$ normalised object units. The reference mechanism constrains the resulting motion. A hinge trial succeeds when its opening reaches $30^\circ$, or $60\%$ of the reference range if that range is below $30^\circ$. A slide trial succeeds at $60\%$ of its reference travel. Success rates include all reference joints, with missing predictions counted as failures. The retry setting allows one attempt in the reverse direction. Oracle trajectories use reference joint parameters.

\begin{table}[htbp]
\centering
\small
\setlength{\tabcolsep}{6pt}
\caption{\textbf{Compliant manipulation on PhysX-Mobility.} Opening success (\%) over all reference joints, including missing predictions as failures. ``Attempted'' denotes prediction coverage (\%). One retry reverses the trajectory direction.}
\label{tab:manip_full}
\begin{tabular}{@{}lr@{\hspace{8pt}}rrr@{\hspace{8pt}}rrr@{}}
\toprule
& & \multicolumn{3}{c}{Open as predicted} & \multicolumn{3}{c}{With one retry} \\
\cmidrule(lr){3-5} \cmidrule(lr){6-8}
Method & Attempted & All & Hinge & Slide & All & Hinge & Slide \\
\midrule
GT joints (oracle) & 100.0 & 99.9 & 100.0 & 99.9 & 99.9 & 100.0 & 99.9 \\
\midrule
PhysX-Anything & 57.2 & 50.3 & 52.1 & 49.6 & 52.3 & 58.0 & 49.9 \\
URDF-Anything+ & 11.3 & 4.4 & 12.8 & 1.0 & 4.9 & 14.0 & 1.2 \\
PhysX-Omni & 56.5 & 48.9 & 51.6 & 47.8 & 51.2 & 58.0 & 48.4 \\
ViewStruct & 91.3 & \textbf{74.6} & \textbf{56.7} & \textbf{81.8} & \textbf{80.3} & \textbf{74.1} & \textbf{82.7} \\
\bottomrule
\end{tabular}
\end{table}

\paragraph{Results and failure analysis.}
As shown in \Cref{tab:manip_full}, direction reversal increases ViewStruct's hinge success from $56.7\%$ to $74.1\%$. Among $217$ initial hinge failures, $169$ exhibit no motion; the reverse-direction retry recovers $94$ failures. These results highlight a limitation of sign-invariant axis metrics: successful execution also depends on the motion direction specified by the axis and joint interval. Pivot errors can further cause the commanded trajectory to conflict with the reference mechanism.

\subsection{Panda Manipulation in GRScenes}

We evaluate a Panda arm in MuJoCo over ten trials in ten GRScenes rooms. The arm follows trajectories derived from ViewStruct's predictions, while each object retains its reference mechanism. A compliant point constraint provides assisted contact, robot--scene collisions are disabled, and foreground geometry is cut away for visibility. Eight trials reach the opening threshold. This evaluation assesses trajectory execution under assisted contact; it does not evaluate collision-free planning or autonomous grasp acquisition.

\begin{figure}[t]
    \centering
    \includegraphics[width=\linewidth]{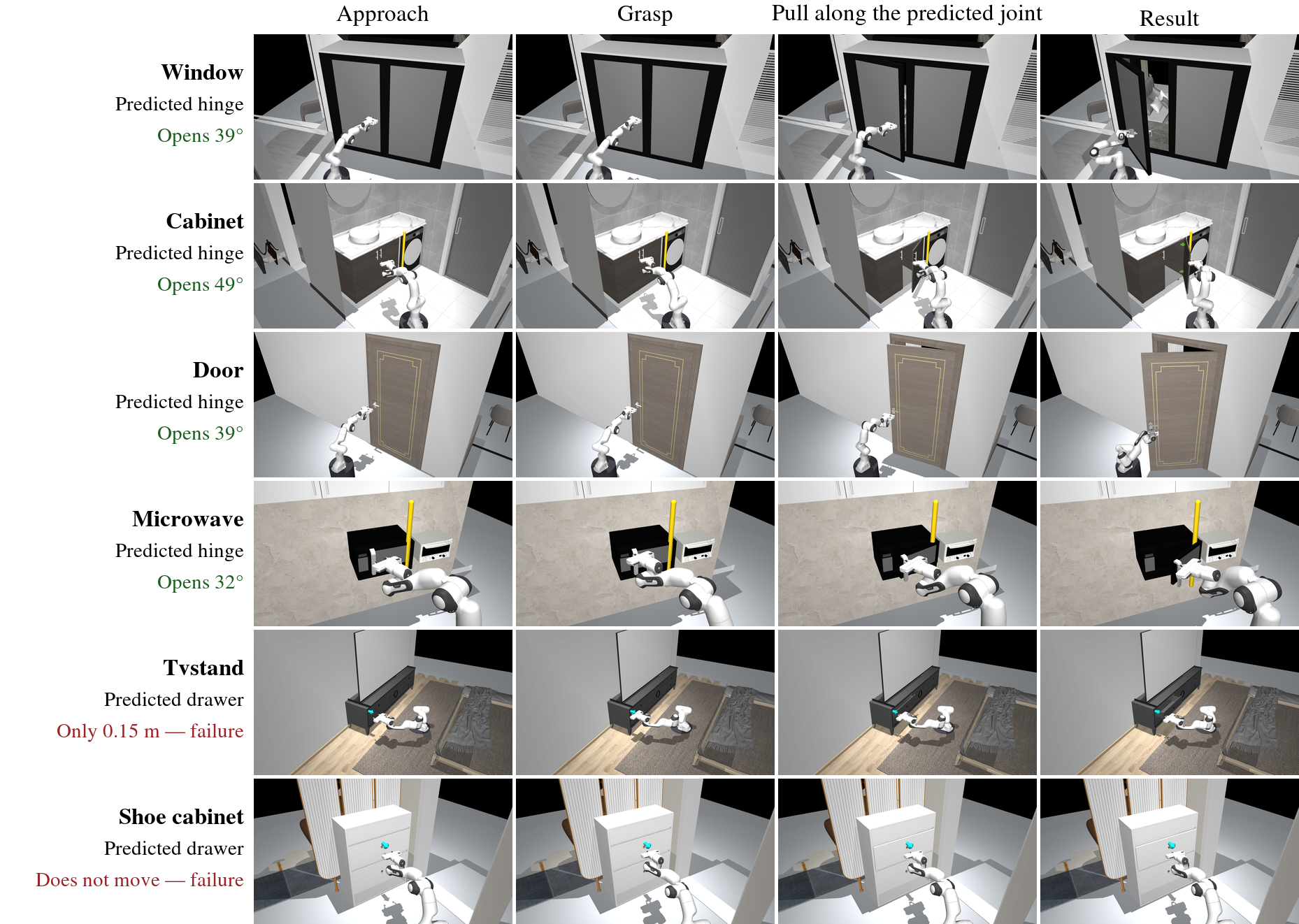}
    \caption{\textbf{Additional Panda manipulation trials.} Four successful openings and two failures under reference mechanisms with assisted grasps. Outcomes are discussed in the text.}
    \label{fig:mujoco_supp}
\end{figure}

\Cref{fig:mujoco_supp} shows successful openings of a window ($39^\circ$), cabinet door ($49^\circ$), room door ($40^\circ$), and microwave door ($32^\circ$). In the two unsuccessful trials, the TV-stand drawer stops $3$\,cm short of its threshold, while the shoe cabinet remains stationary because the predicted sliding direction points into the reference joint limit. These cases illustrate how insufficient travel and incorrect motion direction can prevent opening even when the unsigned axis is plausible.

\section{Limitations and Future Work}
\label{sec:supp_limits}

\paragraph{Input assumptions and observational ambiguity.}
ViewStruct relies on a supplied part decomposition: missing parts or regions spanning independently moving bodies can prevent correct structural recovery. Geometric accuracy also depends on depth, camera calibration, gravity estimation, and consistent part association across observations. 
Our current image-based experiments additionally supply the joint count; estimating it jointly with structure would relax this assumption. 
Even with accurate inputs, static observations may not uniquely determine hidden pivots, opening directions, or mechanical limits. Additional views and motion evidence can reduce these ambiguities, motivating future work on uncertainty estimation and active observation selection.

\paragraph{Model capacity and training coverage.}
Computational constraints restrict our experiments to a 4B-parameter VLM and a finite collection of articulated assets. Although the model transfers to unseen categories and real-world interactions, its task-specific training uses exclusively synthetic data, leaving a domain gap in appearance, segmentation quality, and depth accuracy. Larger backbones and more diverse training observations may improve robustness, although their benefits must be balanced against inference cost.

\paragraph{Closed-loop robotic interaction.}
Inference takes a few seconds, making ViewStruct suitable for initial structure estimation and high-level planning, while faster updates would be needed for continuous feedback control. Our manipulation evaluation uses supplied contact locations and simplified physics, and does not assess autonomous grasp selection, collision avoidance, or physical-property estimation. Future work could initialise an interaction loop with ViewStruct, track part motion during contact, and refine joint parameters before replanning. Geometric fitting could provide these updates without retraining the predictor.

\end{document}